\documentclass[11pt]{article}
\usepackage{acl}   %
\usepackage{times}
\usepackage{latexsym}
\usepackage[T1]{fontenc}
\usepackage[utf8]{inputenc}
\usepackage{microtype}
\usepackage{inconsolata}
\usepackage{graphicx}
\usepackage{amsmath,amssymb}
\usepackage{booktabs}
\usepackage{multirow}
\usepackage{enumitem}
\usepackage{xcolor}
\usepackage{hyperref}
\graphicspath{{figs/}}
\newcommand{\arm}[1]{\textsc{#1}}

\title{OraclePhys: A Systematic Framework for\\
LLM Fine-Tuning on Structural Mechanics}
\author{
  \textbf{Mingyu Li\textsuperscript{1}},
  \textbf{Guorui Song\textsuperscript{2}},
  \textbf{Jing Lin\textsuperscript{2}},
  \textbf{Haoqian Wang\textsuperscript{2}}
\\
\\
  \textsuperscript{1}University of Houston,
  \textsuperscript{2}Tsinghua University
\\
  \small{
    \textbf{Correspondence:} \href{mailto:mli52@cougarnet.uh.edu}{mli52@cougarnet.uh.edu}
  }
}

\begin{document}
\maketitle
\begin{abstract}
What fine-tuning teaches a model is usually diagnosed after the fact.
We make it an experimental variable. \textbf{OraclePhys} has three parts. \textbf{OraclePhys-Bench} is a structural-mechanics benchmark whose finite-element oracle grades every answer and counterfactual edit. \textbf{OraclePhys-30K} holds seven answer forms over byte-identical structure descriptions. A controlled \textbf{training study} compares the forms and verifier roles. Two findings follow. First, within this
task family the answer form of the label determines what fine-tuning
teaches. Ranking supervision lifts held-out localization of the most-deforming story on taller structures from 0.25 to 0.86, a peak-value label yields at best a partial predictor, and a pass/fail label none. This ranking-versus-peak gap replicates on a second domain, a second model family and a paraphrased surface. Second,
the gap between our standard SFT and GRPO runs is mostly a recipe
effect. With learning rate and step budget matched, the two reach held-out accuracies within 0.03 of each other. At higher rates, GRPO falls below its start while SFT does not. Both also teach the model to apply edits given as instructions. The trained 8B matches a specialist and exceeds a frontier LLM. Both domains are linear,
steady-state systems with synthetic text inputs.
\end{abstract}

\section{Introduction}\label{sec:intro}
Training objectives are usually chosen for gradeability: a scalar loss, a
pass/fail check, a preference score. What a language model
\emph{internalizes} under a given objective is diagnosed only after the
fact, by probing or benchmarking---a literature of striking negative results \citep{vafa2024world,wang2024simulator} built on observation rather than intervention. Whether the objective itself causally determines
\emph{what} is internalized, everything else held fixed, remains to be isolated as an experimental variable.

\begin{figure*}[t]
\centering
\includegraphics[width=0.85\textwidth]{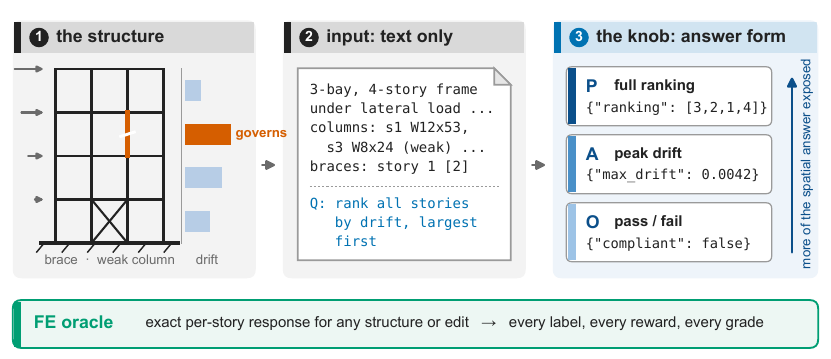}
\caption{The testbed. A steel frame is described in natural language only, an FE oracle grades every answer and every counterfactual edit, and the knob sets how much of the oracle's spatial answer the objective exposes (Eq.~\ref{eq:knob}).}
\label{fig:overview}
\end{figure*}

We isolate it with \textbf{OraclePhys-Bench} (\S\ref{sec:testbed}): a
finite-element solver returns the exact per-story lateral response of
any steel frame.

The testbed is a \emph{model organism} for a question about training rather than a benchmark of engineering practice. Whether the label's answer form
determines what fine-tuning teaches matters well beyond this domain
\citep{kung2023models,webson2022prompt,zhou2023lima}. Testing it requires byte-identical inputs, controlled labels, and a
grader that scores every counterfactual---properties this domain provides by construction and natural corpora lack \citep{li2023othello}. The cost is that all descriptions follow one template (Limitations).

The experiment trains on \textbf{OraclePhys-30K}, holding the base
model, training structures, and input text byte-identical and varying just the answer form the objective demands:
full per-story ranking (\arm{P}), peak scalar (\arm{A}), or pass/fail
boolean (\arm{O}). A model trained on full rankings localizes the \emph{governing story}, the one that deforms most, of held-out taller structures at 0.86 or better, from a base near chance. The scalar label yields at best a partial predictor, and the boolean none
detectable (\S\ref{sec:staircase}). The gap between ranking and scalar supervision
replicates on a second physics domain, a second model family, and a
paraphrased evaluation surface.

The same oracle then enters training in two ways, as written answers
(SFT) or as scores of sampled answers (GRPO)
(\S\ref{sec:verifier}--\ref{sec:routing}). At their standard recipes,
answers raise held-out accuracy and scores do not, the familiar pattern
in which SFT beats RL. Matching learning rate and step budget from one
start removes most of that gap. At higher rates the two diverge: GRPO falls below its start while SFT does not. Until the
recipe is matched, an SFT-versus-RL comparison is a comparison of
recipes. SFT and GRPO alike also teach the model to apply an edit stated as an action, a skill it lacked even when it could predict the edited structure.

Reference points (\S\ref{sec:ceilings}) calibrate these numbers: the
trained 8B matches a specialist GNN and exceeds a frontier model that sits near a simple 30-line physics formula.

\noindent\textbf{Contributions.}
\begin{itemize}[leftmargin=1.1em,labelsep=0.5em,itemsep=2pt,topsep=2pt,parsep=0pt]
\item We build \textbf{OraclePhys-Bench}, an exactly-graded testbed for
structural mechanics, with verified surface-cue reduction, intervention
probes, and an isomorphic second domain.
\item We release \textbf{OraclePhys-30K}, a supervision dataset of seven
answer forms over byte-identical structure descriptions.
\item We train an 8B LLM on per-story structural response in a
controlled study (seven supervision forms, three verifier roles, one
exact oracle). It scores above Claude Opus~4.8 at zero- and 32-shot and matches a specialist GNN.
\item We establish two \textbf{findings}. Within the task families
tested, the answer form of the label determines what fine-tuning
teaches, and which form ranks highest depends on the readout
(\S\ref{sec:staircase}). The gap between our standard SFT and GRPO runs
is mostly a recipe effect: with learning rate and step budget matched, most of it
disappears (\S\ref{sec:verifier}--\ref{sec:routing}).
\end{itemize}

\section{An Oracle-Graded Testbed for Spatial Forward Models}\label{sec:testbed}

\subsection{Task and oracle}\label{sec:task}
The subject domain is the lateral response of 2-D steel frames
(Fig.~\ref{fig:overview}): multi-story, multi-bay, with per-story
sections, a connection type, optional braces, damage events, and a
lateral load whose vertical profile varies. An OpenSees finite-element
model returns, for any structure, the \emph{per-story inter-story drift
ratios}
\begin{equation}
\delta_k \;=\; \frac{u_k - u_{k-1}}{h_k},
\qquad
k^{*} \;=\; \arg\max_{k}\,\delta_k,
\label{eq:drift}
\end{equation}
where $u_k$ is the lateral displacement of floor $k$ ($u_0{=}0$ at the
base) and $h_k$ the story height. The \emph{governing story} $k^{*}$ is
the one that governs the design. Soft or damaged stories drift
more, and a brace stiffens its story. The
FE solution \emph{defines} ground truth for this task, and grading against it
is deterministic and exact. Every label and every grade comes from the solver.

The model reads a \emph{natural-language description} (geometry, sections, connections, braces, damage, load), its sole input, and must rank all stories by drift, in JSON. Primary metrics: \textbf{top-1 localization}
of the governing story and Spearman $\rho$ of the full ranking, on
held-out structures ($n{=}150$ per axis, 80 for extrapolation).

Two properties make this testbed diagnostic rather than merely hard.
Structural weakness is \textbf{spatially non-local}: identifying the governing story requires combining several sentences of the description. And every claimed
capability can be \textbf{probed by intervention}, because the oracle
grades counterfactuals: ``if a story is strengthened, where does the weakness move?'' has a verified answer. We use two terms in a narrow sense. ``Forward
model'' refers only to performance on the intervention-probing axes of
\S\ref{sec:axes}, a \emph{behavioral} description rather than a claim about internal representations. ``Routing'' names one measured pattern
(\S\ref{sec:routing}): action-application accuracy rises while held-out
ranking accuracy does not. 

\subsection{Evaluation axes}\label{sec:axes}
All axes use deterministic instance streams (\S\ref{sec:stats}) and the same
parser. A parse failure scores as a miss.

\textbf{Forward localization / ranking}: rank the stories of an unseen,
taller structure (\S\ref{sec:stats}).
\textbf{Post-intervention forward}: the governing story is strengthened, and the model must localize the \emph{new} one from the edited description---anti-correlated
with the pre-edit answer by construction (edited structures of this kind appear in training, see Limitations).
\textbf{Two-step horizon}: two sequential edits---errors compound.
\textbf{Extrapolation}: forward localization on the tallest structures.
\textbf{Action interface}: the \emph{original} structure plus a
one-sentence action (``add a brace to story 3''), applied internally. A
paired control rewrites the same transition as a full post-edit
description, so the act$-$desc gap isolates action-application
(\S\ref{sec:routing}). \textbf{Invariance} (sanity): a strengthened story must stay at least as strong relative to the others.

\subsection{Two difficulty tiers, and removing the surface cue}\label{sec:tiers}
\textbf{Tier-1 (realistic)} structures follow design practice, leaving exploitable regularities. \textbf{Tier-2 (hardened)} randomizes them
(section jitter, composed weakening, varied load shapes). Hardening is
\emph{verified}: a bag-of-words probe falls from 0.63 to near chance, a
structured-feature probe loses a third of its margin over chance, and a geometry-only
probe stays near its tier-1 value (Appendix~\ref{app:testbed}).

The non-learning reference is a \textbf{30-line first-order formula}
(each story's shear force divided by a proxy for its stiffness). It is a deliberately strong baseline, since it reads numeric section properties directly from the design
object, and its tier-2 scores sit well above random (Table~\ref{tab:d2}): any
physics claim must clear this row.

\subsection{Splits, determinism, and statistics}\label{sec:stats}
Training structures have 3--4 stories and 2--3 bays, and every held-out
structure has 5--6 stories and 3--4 bays, so every test ranking is longer
than any seen in training (chance 0.18 for top-1). In
the taxonomy of \citet{hupkes2023genbench} this is a covariate shift in
structure size: size generalization within one task family rather than a topological shift.
All streams generate deterministically from fixed seeds: every run,
across models, checkpoints and months, answers the \emph{same} 150
questions per axis.
This gives \textbf{bit-exact reproducibility} under greedy decoding and
fixed batching, and \textbf{paired statistics}: per-instance correctness
arrays are stored in every results file, and all model comparisons are
exact McNemar tests \citep{mcnemar1947} on discordant pairs with paired
bootstrap CIs. Of the ${\sim}50$ paired tests reported, the staircase claims rest on
$p<10^{-4}$ effects and the \S\ref{sec:verifier} contrasts on powered
$n{=}600$ streams (Table~\ref{tab:d7}). Results that clear the uncorrected threshold but miss the FDR correction of Table~\ref{tab:d4} count as secondary evidence. Provenance is \textbf{machine-checkable}: one script regenerates every number from archived per-instance files. The exported streams, grader and reference rows constitute \textbf{OraclePhys-Bench}, and new streams can be generated on demand to guard against contamination.

\section{The Objective-Shape Staircase}\label{sec:staircase}

\subsection{One knob: how much of the answer the objective carries}
Three training arms share the base model (Qwen3-8B + LoRA), update
budget, ${\sim}4$k training structures, and \textbf{byte-identical input
text} (verified by a token-level leakage audit). They differ in the question-and-answer suffix alone. With
$\boldsymbol{\delta}$ from Eq.~\ref{eq:drift} and drift limit $\tau$,
the three labels are
\begin{equation}
\begin{aligned}
y_{\mathrm{P}} &= \operatorname{argsort}_{\downarrow}\,\boldsymbol{\delta},
\qquad
y_{\mathrm{A}} = \max_k \delta_k,\\[2pt]
y_{\mathrm{O}} &= \mathbf{1}\!\left[\max_k \delta_k \le \tau\right],
\end{aligned}
\label{eq:knob}
\end{equation}
so that \arm{P} (\textbf{ranking}) names the relative position of every
story, \arm{A} (\textbf{scalar}) names a single number that depends on
the governing story but does not name it, and \arm{O} (\textbf{boolean}
compliance) names one of two outcomes.
Arm \arm{A} models a common failure: getting the peak value right while missing which story it belongs to (\S\ref{sec:format}). The knob is \textbf{how much of the oracle's spatial answer the objective exposes}, and we call the resulting ordering of arms the \emph{staircase}. OraclePhys-30K
totals ${\sim}30$K oracle-graded examples across the seven arms
(${\sim}3.8$k rows each, Table~\ref{tab:arms}), the hardened second
round, and the action pairs (Appendix~\ref{app:training}).
\begin{table}[t]\centering
\caption{The seven arms: tier-1 governing-story localization in each
arm's own format without examples (\emph{raw}; 3-seed mean for \arm{P},
\arm{A}, \arm{O}), with two worked examples prepended (\emph{calibrated};
three seeds, mean${\pm}$SD), and on the story-naming arms' own trained
output (\emph{own readout}). All axes: Tables~\ref{tab:d1},
\ref{tab:readout}.}
\label{tab:arms}\vspace{1ex}\scriptsize\setlength{\tabcolsep}{3pt}
\begin{tabular}{llcccc}
\toprule
 & & & \multicolumn{2}{c}{calibrated} & own \\
\cmidrule(lr){4-5}
arm & label names & raw fwd & fwd & post-int. & readout \\
\midrule
\arm{P} & full drift ranking & 0.86 & 0.74${\pm}$0.01 & 0.74${\pm}$0.04 & 0.89 \\
\arm{G} & governing story only & 0.85 & 0.75${\pm}$0.07 & 0.69${\pm}$0.02 & 0.88 \\
\arm{V} & per-story drift values & 0.52 & 0.64${\pm}$0.03 & 0.39${\pm}$0.08 & 0.89 \\
\arm{K} & one story's drift & 0.56 & 0.56${\pm}$0.07 & 0.36${\pm}$0.01 & -- \\
\arm{A} & peak drift (one scalar) & 0.56 & 0.50${\pm}$0.09 & 0.47${\pm}$0.04 & -- \\
\arm{O} & pass/fail compliance & 0.15 & 0.32${\pm}$0.05 & 0.18${\pm}$0.01 & -- \\
\arm{S} & shuffled ranking (placebo) & 0.24 & 0.37${\pm}$0.03 & 0.11${\pm}$0.01 & -- \\
base & (untrained) & 0.25 & 0.30 & 0.16 & -- \\
\bottomrule\end{tabular}\end{table}

\subsection{The ranking--scalar gap, three times}
\begin{figure}[t]
\centering
\includegraphics[width=\columnwidth]{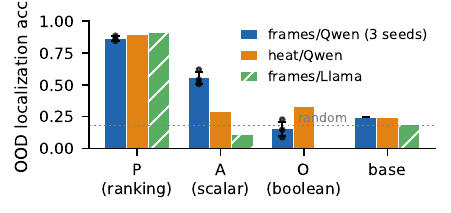}
\caption{The ranking--scalar gap in three settings (frames with Qwen over 3 seeds, heat with Qwen, frames with Llama). Bars show OOD localization and the dotted line random.}
\label{fig:staircase}
\end{figure}

Fig.~\ref{fig:staircase} shows OOD governing-story localization in three
settings: frames/Qwen (3 seeds), heat/Qwen, and frames/Llama. \arm{P}
localizes the governing story in all three, at 0.86 or better. \arm{A}
and \arm{O} separate cleanly from the base in the primary setting
(Appendix~\ref{app:tables}). On heat, \arm{A} and \arm{O} score the same, and the Llama setting omits the \arm{O} arm. The second domain is 2-D steady-state heat
conduction, a structural isomorph (rows $\leftrightarrow$ stories,
cooling $\leftrightarrow$ braces) with its own surface-cue audit. The scalar arm's partial gain is specific to frames with Qwen. It is absent on heat and, under calibration, on Llama (Tables~\ref{tab:d1}, \ref{tab:d3}). The
ranking--scalar gap is the invariant, and the full ordering holds in the
primary setting under paired tests (\S\ref{sec:format}). The two domains
share no exploitable surface: the frames-trained \arm{P} arm, transferred
zero-shot to heat, collapses to chance-level localization while its output format survives intact.

\subsection{Format-calibrated control: knowledge, not output format}
\label{sec:format}
\arm{A} and \arm{O} never produced a ranking in training, so the raw
staircase conflates missing knowledge with format failure. In the raw protocol two trained cells score below the untrained base, a sign of a measurement artifact rather than lost knowledge. We
therefore re-ran every arm with two worked examples prepended (the \emph{calibrated} protocol, used on the tier-1 stream in \S\ref{sec:format}--\ref{sec:ladder} and labelled \emph{fmt2} in the appendix tables). The
prompt is the same for every arm, so it equalizes format while adding
little task information (the untrained base barely moves,
Table~\ref{tab:arms}, cf.\ \citealp{min2022rethinking}).

Under calibration the staircase stands across three seeds
(Table~\ref{tab:arms}): \arm{P} leads \arm{A} by a wide margin, \arm{A}
leads the base, and \arm{O} sits at the base. Both leads are significant
on the forward and on the post-intervention axis, and \arm{O} stays at the base (Tables~\ref{tab:d4} and~\ref{tab:d5}). Both below-base cells disappear, as an artifact would predict. On Llama the scalar arm stays at the base, and the ranking--scalar gap is wider than on Qwen. \textbf{The
ranking objective yields the full spatial ordering on held-out
structures, the scalar objective a real but partial one, and the boolean
objective none detectable.} What the label \emph{spells out} about the spatial answer, in directly usable form, is what fine-tuning teaches. The first case of Table~\ref{tab:cases} shows the difference. Story 3 has the lightest columns but is braced, and story 5, the only unbraced story, governs. The ranking arm reproduces the oracle's ranking exactly. The scalar arm and the untrained base both put story 3 first, following the lightest columns, and the boolean arm returns the stories in input order.
Calibrated \arm{A} stays below the first-order formula, so its partial gain is weak. \arm{P} itself dips under prepended
examples, which are off-distribution for an SFT'd model. The raw and calibrated protocols order the arms identically.

\paragraph{The staircase survives a second linguistic surface.}
We re-rendered the held-out instances through a paraphrased template
(running prose and reordered fields, with identical information and parser) and
re-evaluated the \emph{frozen} checkpoints. The staircase is unchanged: every calibrated arm's forward accuracy stays within 0.03 (seed-42 checkpoints), and raw \arm{P} loses little
(Table~\ref{tab:d8}). What the ranking objective
teaches is readable from an unseen surface.

\subsection{Controls: format, task proximity, and answer content}
\label{sec:ladder}
Does the staircase measure the label's content, or merely the proximity
between training task and evaluation? Four further arms
(Table~\ref{tab:arms}), all on byte-identical structure text, separate
the candidates. \arm{K} keeps
\arm{A}'s scalar format but changes the content: each example asks for the
drift of one randomly designated story, exposing the per-story field one story at a time. \arm{V} asks for the full drift-value
vector, from which the ranking follows. \arm{G} asks just \emph{which} story governs. \arm{S} is the placebo: \arm{P}'s format verbatim with
shuffled permutations as labels, a training-side control task
\citep{hewitt2019control}.

\textbf{Three seeds for every arm.} Table~\ref{tab:arms} gives the
calibrated forward and post-intervention scores as three-seed means.
\arm{P} $>$ \arm{A} $>$ \arm{O} holds across seeds, with \arm{P} ahead
of \arm{A} by about 0.25 on both axes. \arm{G} ties \arm{P}, and \arm{K} ties \arm{V}.

\textbf{The readout matrix.} The axes above ask ordinal questions, which favor the ranking arm's trained output. We
therefore scored every arm under six readouts, one per trained target, plus the argmax of the value readout,
on all four axes, with and without two worked examples
(Table~\ref{tab:readout}). Read on its own trained output, each
story-naming arm localizes equally well (last column of
Table~\ref{tab:arms}), and \arm{P}, \arm{G} and \arm{V} stay together on
the other three axes. \arm{A}'s best ordinal readout stays below them on
every axis, and \arm{O} and \arm{S} stay near the base on every ordinal
readout (cases in Table~\ref{tab:cases}). Each arm learns its own target: \arm{O} beats the
majority baseline on held-out compliance while its ordinal readouts stay
at base. Numerical readouts improve on the base but stay coarse, so
\arm{V}'s ordering is learned better than its magnitudes.

Three readings follow. \emph{Format alone teaches nothing}: the placebo
stays near the base and degrades intervention behavior. \emph{The
standing of supervision forms depends on the readout}: under the ranking readout \arm{P} and \arm{G} lead, \arm{V} and \arm{K}
follow, and \arm{A} trails (Tables~\ref{tab:arms} and~\ref{tab:readout}),
while under each arm's own readout the story-naming arms coincide. Value supervision also supports localization: the answer form
determines which usable function is learned and under which readout it
shows. \emph{Proximity is a partial explanation}: \arm{V} and \arm{G},
never trained to rank, match \arm{P} under their own readouts. Because \arm{K} differs from \arm{P} in coverage and content as well as proximity, the \arm{P}--\arm{K} contrast mixes the three.

\paragraph{What emerges without being trained.}\label{sec:emergence}
Training contains single-structure ranking questions only. Yet the
\arm{P} arm acquires two-step rollout, taller-structure generalization
from 3--4-story supervision, and post-intervention localization, each
far above the base, and on heat the same untrained axes emerge
(Tables~\ref{tab:d1}, \ref{tab:d3}). \arm{A}, with identical data and
compute, acquires them in attenuated form. The two-step and extrapolation axes require more than replaying training statistics. 

\section{One Oracle, Two Uses: Matching the Recipe Closes Most of the SFT--GRPO Gap}\label{sec:verifier}

\begin{table*}[t]\centering
\caption{One start, two uses of the oracle: every row trains from the
tier-1 \arm{P} arm on the same hardened pool (seed 42). Tier-2 forward
and post-intervention localization ($n{=}150$), the forward gain over
the start (exact McNemar $p$), accuracy on fresh $n{=}600$ streams pooled
over three tier-2 axes (1800 instances; Table~\ref{tab:d7}), and tier-1
forward retention. The matched cells share GRPO round-2's budget (600
steps, no reuse, $g{=}16$, $\beta{=}0.01$); per-axis cells in
Table~\ref{tab:matched}.}
\label{tab:twouses}\vspace{1ex}\scriptsize
\begin{tabular}{lccccc}
\toprule
cell & fwd & post-int. & $\Delta$fwd vs.\ start ($p$) & pooled $n{=}600$ & tier-1 fwd \\
\midrule
start: the \arm{P} arm & 0.660 & 0.553 & -- & 0.622 & 0.887 \\
dense SFT (answers; $10^{-4}$, 2 epochs) & 0.747 & 0.693 & +0.087 (0.03) & 0.716 & 0.867 \\
RFT (filtered answers; dense recipe) & 0.753 & 0.680 & +0.093 (4e-3) & 0.710 & 0.820 \\
GRPO round-2 (scores; $10^{-6}$, 600 steps) & 0.687 & 0.633 & +0.027 (0.42) & 0.629 & -- \\
\midrule
matched SFT, $10^{-6}$ & 0.667 & 0.567 & +0.007 (1.00) & 0.637 & 0.893 \\
matched SFT, $10^{-5}$ & 0.760 & 0.660 & +0.100 (3e-4) & 0.666 & 0.880 \\
matched SFT, $3{\times}10^{-5}$ & 0.767 & 0.640 & +0.107 (4e-3) & 0.672 & 0.873 \\
matched SFT, $10^{-4}$ & 0.687 & 0.667 & +0.027 (0.63) & 0.662 & 0.827 \\
matched GRPO, $10^{-5}$ & 0.760 & 0.667 & +0.100 (0.01) & 0.640 & 0.720 \\
matched GRPO, $3{\times}10^{-5}$ & 0.627 & 0.600 & -0.033 (0.55) & 0.571 & 0.533 \\
matched GRPO, $10^{-4}$ & 0.487 & 0.367 & -0.173 (2e-4) & 0.434 & 0.407 \\
\bottomrule\end{tabular}
\par\vspace{2pt}\parbox{\textwidth}{\scriptsize Three-seed means (mean${\pm}$SD) where seeds exist: dense SFT forward $0.731{\pm}0.011$, post-intervention $0.691{\pm}0.008$ (each seed significant over the start, $p\leq0.0013$); best-practice GRPO forward $0.751{\pm}0.030$ against its 0.747 start; lower-start GRPO forward $0.678{\pm}0.014$ against its 0.660 start.}
\end{table*}

\paragraph{Design.}
The oracle enters training in three roles, \emph{teacher}, \emph{filter}
and \emph{judge}, which collapse into two uses: answers versus scores.
Every run in Table~\ref{tab:twouses} starts from the \emph{same}
checkpoint (the tier-1 \arm{P} arm, its first row) and uses the
\emph{same} hardened pool. The
\textbf{teacher} protocol trains on oracle-written rankings (dense SFT), and the \textbf{judge} optimizes oracle scores of sampled rankings (GRPO, budgets in Appendix~\ref{app:training}). Weights, LoRA configuration and
prompts are shared. What varies is \emph{how the oracle's knowledge
enters the gradient} and, in the standard runs, also the learning rate,
step budget and data reuse. The matched cells remove that second difference. A fair test of the update rule needs both methods to see the same prompts for the same number of steps at the same rate, and these cells provide that.

\paragraph{Results.}
At their standard recipes, answers help and scores do not. The teacher
protocol raises every tier-2 axis, significantly in each of three seeds,
with little tier-1 forgetting (Table~\ref{tab:twouses}). GRPO at lr
$10^{-6}$ leaves the forward, extrapolation and two-step axes within
noise \emph{while its own reward climbs from 0.80 to 0.93}
(Tables~\ref{tab:d4},~\ref{tab:d5}), and on the pooled $n{=}600$
streams the dense and filtered endpoints lead it by a wide margin
(Table~\ref{tab:d7}). GRPO also leaves held-out accuracy flat from other starting points. Continued from the dense-SFT checkpoint, it leaves that checkpoint unchanged. From the untrained base, it more than doubles its reward yet localizes at random on the hardened tier (Table~\ref{tab:d1}). The one exception is a boundary case: from the \arm{P} arm, post-intervention accuracy rises by 0.08, but the gain does not survive FDR correction and vanishes on the fresh stream (Table~\ref{tab:d7}).

\paragraph{The matched comparison.}
Matching the recipe removes most of this gap. The standard runs also
differ in steps, learning rate, data reuse, KL coefficient and how
sampled outputs enter the loss (Appendix~\ref{app:training}), so we add
\emph{matched-recipe} cells: same start, same pool, GRPO's step budget
(600 steps, two prompts per step, each prompt used once). At $10^{-6}$ neither update
moves. At $10^{-5}$ both do, and they reach the same forward accuracy on
the shared and on the fresh $n{=}600$ stream. On the pooled streams SFT leads by 0.03, inside the equivalence margin and mostly on the two-step axis. On the pooled streams SFT's gain over the start is significant, and GRPO's is not (Tables~\ref{tab:twouses},~\ref{tab:matched},~\ref{tab:d7}). Above that
rate the two diverge. SFT at $3{\times}10^{-5}$ and $10^{-4}$ stays above its start, while GRPO falls below it and emits runaway
completions (Appendix~\ref{app:training}).
GRPO also keeps less of its tier-1 accuracy, losing a fifth at
$10^{-5}$ and over half at $10^{-4}$, where SFT keeps most of it. The full dense recipe still leads every matched cell. Matched SFT at the dense recipe's own rate ($10^{-4}$) stays below it, so the lead goes with the dense recipe's larger example budget (a bigger batch, a second epoch). On the pooled streams the standard-recipe gap of 0.087 thus splits into 0.050 for the dense recipe (dense SFT over matched SFT at $10^{-5}$), 0.011 for GRPO's learning rate (matched GRPO at $10^{-5}$ over the second round) and 0.026 for the update rule at a matched rate (Table~\ref{tab:twouses}). About seven tenths of the gap is recipe. Appendix~\ref{app:alt} tests five alternative explanations and rules each one out.

\paragraph{Provenance, recipe, and lower-start controls.} Three further controls ask whether the standard-recipe null comes from the pool, the start or the budget. \emph{Filtered self-distillation} (RFT, the STaR/RAFT/ReST recipe,
\citealp{zelikman2022star,dong2023raft,gulcehre2023rest,singh2024rest}):
sample 50 answers per hardened prompt from the \arm{P} checkpoint,
\emph{score} them with the oracle, keep the best with reward $\geq
0.9$, and imitate the survivors. The oracle contributes just scores, yet the gradient is dense imitation. This raises held-out
accuracy on both axes, significantly and indistinguishably from
oracle-written labels (Tables~\ref{tab:twouses},~\ref{tab:d4}), so the hardened pool holds enough learnable signal (the \emph{pool-adequacy control}).
\emph{Best-practice GRPO} uses group size 50, a difficulty-filtered
pool, the strongest SFT base and ${\sim}10\times$ the dense round's
oracle-call budget. Across three seeds it lands where it started (Table~\ref{tab:d4}, seed means in the note to Table~\ref{tab:twouses}).
\emph{Lower-start GRPO} is the identical recipe launched from the
pre-hardening 0.660 start, where the dense round gains $+0.12$ across
the four axes. Across three seeds it does not move on held-out
accuracy, while its in-pool localization climbs (Fig.~\ref{fig:judge}).
\emph{Update-scale probe:} dense SFT trains at ${\sim}100\times$ GRPO's
learning rate with multi-epoch reuse, so we re-ran best-practice GRPO
at tenfold lr ($10^{-5}$) from the dense-SFT checkpoint. It does not
raise held-out accuracy but \emph{lowers} it, on the forward axis and on
the four axes pooled (Table~\ref{tab:d5}). From the lower start, the same rate does raise held-out accuracy
(cf.\ \citealp{xiong2025minimalist,chen2025nft}). The matched cells place the standard-recipe difference in the recipe rather than in token-level credit assignment. The implicit conservatism of on-policy updates \citep{shenfeld2025razor} remains untested. The explicit KL penalty is also ruled out: the second-round, matched and action-format runs used $\beta{=}0.01$,
every other GRPO run $\beta{=}0$, and the standard-rate null is the same
in both groups (Table~\ref{tab:grpo}).

\paragraph{Where the advantage signal vanishes.}
Group-relative RL redistributes probability among behaviors the policy
already samples: with rewards $r_{1:g}$, each advantage is
$\hat{A}_i = r_i - \tfrac{1}{g}\sum_j r_j$, which is near zero on
instances the policy always gets wrong. Among 50 samples per hardened
prompt, 60\% of prompts are stable-correct, 34\% swing and 6\%
stable-wrong, and the tail is shallow (pass@$k$ in
Appendix~\ref{app:alt}). Stable-wrong prompts make up 6\%, too few to explain the null, and the matched cells point to the learning rate instead.

\textbf{Scope.} GRPO with multi-epoch data reuse, full-parameter
training, PPO, DPO, and longer horizons \citep{liu2025prorl} are
untested (Limitations).

\section{The Action Interface: Two Separately Measured Axes}\label{sec:routing}

\paragraph{Knowing the post-edit response without applying an action.}
The dense-SFT model of \S\ref{sec:verifier} predicts the post-edit
response when it is described but not when it must \emph{apply} an
action: handed the original structure plus ``add braces to story 3,'' it
loses 0.15 on both tiers, the action-application gap (Table~\ref{tab:d6}, Fig.~\ref{fig:interface} in the appendix).

Tier-2 localization with the action applied internally (act) and with
the same edit given as a full description (desc):
\begin{center}\footnotesize\setlength{\tabcolsep}{4pt}
\begin{tabular}{lccc}
\toprule
model & act & desc & act$-$desc \\
\midrule
dense SFT (\S\ref{sec:verifier}) & 0.533 & 0.680 & $-$0.147 \\
\, + action-format GRPO (scores) & 0.687 & 0.687 & $+$0.000 \\
\, + action-format SFT (answers) & 0.760 & 0.720 & $+$0.040 \\
\bottomrule
\end{tabular}
\end{center}
\vspace{-0.6ex}

\paragraph{Scores close the action gap.}
Action-format GRPO (state--action prompts, rewarded by the oracle's score of the predicted post-action ranking) closes the gap exactly: on both tiers
action-application accuracy rises to the level of the description
control, and the forward and post-intervention axes stay at parity
(Tables~\ref{tab:d4}, \ref{tab:d6}). This is the pattern we call
\emph{routing} (\S\ref{sec:task}). The model already gets about half of the interface task right, so its samples contain many successes for RL to reinforce. The run
used lr $10^{-6}$ (Table~\ref{tab:grpo}), the rate at which held-out accuracy stayed flat under both updates in \S\ref{sec:verifier}.

\paragraph{Routing is shared by RL and SFT: the dense control.} If action-application accuracy just takes training \emph{on} the
interface, dense supervision should raise it too, and it does.
SFT on the same action distribution (4k oracle-labeled pairs) closes
the gap just as completely, with no act$-$desc gap of its own and no
cost on the forward axis (Table~\ref{tab:d6}). On tier-1 it \emph{beats}
the GRPO variant on post-intervention localization by 0.14 and reaches
the paper's highest post-intervention score (Tables~\ref{tab:d1},
\ref{tab:d5}).

\paragraph{Two axes, measured separately.}
Training on the action interface raises action-application accuracy to
the level of the description control, whether the oracle scores sampled
answers (GRPO) or writes them (SFT). On the interface, only oracle-written answers also raise held-out ranking accuracy (action-GRPO ran at $10^{-6}$ only). Action-application and held-out accuracy are thus two separate axes, each reachable by more than one method. Practically, pure SFT
yields the best all-around model (action-SFT). The comparisons in \S\ref{sec:ceilings} use the final assembled checkpoint.

\paragraph{The thinking channel.}\label{sec:think}
An explicit reasoning channel adds nothing in three pre-specified
tier-2 contrasts, and restoring the channel after direct-answer SFT has disabled it costs held-out accuracy (Appendix~\ref{app:think}, Table~\ref{tab:d4}).

\section{Reference Points: Frontier Model, Specialist, and the Convergence Band}\label{sec:ceilings}

\paragraph{The frontier model we test sits near a 30-line formula.}
Under a controlled API protocol (Appendix~\ref{app:training}), Claude
Opus 4.8 localizes the governing story on half of the hardened
instances (Tables~\ref{tab:d1}, \ref{tab:d2}). Against the first-order
formula of \S\ref{sec:tiers} its largest lead is 0.10, on tier-1 forward. Elsewhere its lead is at most secondary evidence (Tables~\ref{tab:d4}, \ref{tab:d5}).

\paragraph{In-context examples leave the gap open.}
The comparison is asymmetric: our model saw 4k oracle-labeled
structures, the frontier model zero. The in-context control narrows it.
With 32 oracle-labeled worked examples Opus stays statistically at the formula (Table~\ref{tab:d5}), while the trained 8B
exceeds this 32-shot frontier by 0.30 (Table~\ref{tab:d4}, Fig.~\ref{fig:frontier} in the appendix). The remaining gap therefore measures what 4k training structures add beyond 32 worked examples in context. \paragraph{A specialist baseline at a similar level.}
A per-story GNN reading \emph{structured graph input} (sections, bracing,
damage, load shape) trains on the same distribution and evaluates on
the same instance streams. On tier-2 it
matches our final text-reading model exactly on the forward and on the
post-intervention axis (Table~\ref{tab:d4}). On fresh streams the pooled
difference is bounded within $\pm0.06$, with the point estimate toward
the LLM (Table~\ref{tab:d7}). On tier-1 the LLM leads on both axes, 0.84 against 0.82 forward and 0.73 against 0.63 post-intervention, within noise (Tables~\ref{tab:d1}, \ref{tab:d5}).

\paragraph{The 0.73--0.78 convergence.}
Six learners with different inductive biases converge on the same
tier-2 band, 0.73--0.78, pairwise indistinguishable (Appendix~\ref{app:tables}).
Fig.~\ref{fig:wall} is a sensitivity illustration: a third of tier-2
instances have a top-2 drift margin under 10\%, so a small hypothetical
per-story error would cap top-1 near the observed band if near-ties were
scored at chance. It is not an upper bound:
the task is noiseless. Shared data limitations, optimization choices,
common model bias, discretization and input precision are untested
alternatives.

\paragraph{What moves the band.}
On the data-efficiency curve (Fig.~\ref{fig:ndata}), tier-1 saturates
early, and tier-2 gains little from the last doubling of data, while one difficulty-matched second round adds 0.09, as much as three doublings. The limiting factor is training data near the decision boundary, where stories nearly tie.

\section{Related Work}\label{sec:related}
\textbf{World models in language models.} Probing shows sequence models
can carry structured internal state
\citep{li2023othello,nanda2023emergent,kim2023entity}, and audits show good
prediction can coexist with a poor model of the generating process
\citep{vafa2024world,wang2024simulator,vafa2025inductive}. These studies observe, whereas we intervene. Physics-of-LMs \citep{allenzhu2024physics}
manipulates data properties, while we manipulate the answer form with the data frozen. Supervision-content manipulations exist---traces with answers fixed
\citep{bhambri2026traces}, formats in logic tuning
\citep{zhou2025dissecting}---without byte-identical inputs or
counterfactual grading. \citet{prakash2024finetuning} argue fine-tuning
enhances existing mechanisms, whereas our base starts near chance (Table~\ref{tab:arms}).

\textbf{Limits of verifier-scored RL.} Whether RLVR expands or merely
elicits capability is contested
\citep{yue2025pass,chu2025sft,liu2025prorl,shao2024deepseekmath,guo2025r1,shao2025spurious}, and curation methods now \emph{assume} RL needs the right SFT
seed---the premise we measure \citep{yao2026tailored}.
\citet{gan2026lrgated} report GRPO degrading at a middle learning rate
and collapsing at a high one, with no gain over a strong supervised baseline. We add the matched comparison: SFT and GRPO from one start, on one pool and one step budget, at each rate. We contribute an oracle-graded controlled cell (one verifier, one
start) and a positive characterization: scores raise action-application
accuracy, and on held-out accuracy the difference between the two comes from the recipe rather than the update rule.

\textbf{LLMs for structural engineering.} In zero-shot
translator-plus-solver systems \citep{masse2025} the LLM translates while a hardcoded
solver computes, and spatial-physics benchmarks test the LLM alone \citep{sphyr2025}. \citet{beamperl2026} RL-tunes a compact LLM on scalar
beam statics (24 test instances, no SFT or control arm) and also sees reward and robustness come apart. Ours is, to our knowledge,
the first controlled study of an LLM trained on \emph{spatial}
structural response.

\section{Conclusion}\label{sec:conclusion}
OraclePhys turns ``what did training teach?''\ into a controlled
variable. Within the task families tested, the label's answer form
determines what fine-tuning teaches, and the readout determines what is
visible. Here the SFT-versus-RL comparison turned out to be mostly a
comparison of recipes: matched in learning rate and step budget, SFT and
GRPO reach nearly the same held-out accuracy, and at higher rates GRPO
falls below its start while SFT does not. Scores do teach a model
to apply edits it could already predict.

\textbf{Implications for ML and NLP.} Three results carry beyond the physics. When a verifier exists, the form of the answer it grades
is a training variable in its own right: one oracle produced seven
supervision forms from one input, and the form decided what the model
could do afterwards and under which readout it showed. Studies of
supervision content in NLP (instruction formats, reasoning traces,
process versus outcome labels, \citealp{webson2022prompt,bhambri2026traces,uesato2022process,lightman2023verify}) can use
the same design wherever a grader for counterfactuals exists. Code with unit tests, mathematics with checkable answers and tool use with verifiable outcomes all admit labels that range from a pass/fail bit to a full worked target. Here the cheapest of these labels, pass/fail, taught the least, and the design above tests whether that holds in those tasks. A standard-recipe SFT-versus-RL gap says little about the update rule until a matched-recipe cell has been run. Here most of the gap came from the recipe. Reward can also rise while held-out accuracy stays flat (\S\ref{sec:verifier}), so both should be reported.

\section*{Limitations}
\textbf{One scale, one recipe, synthetic domains.} All results are
8B~+~LoRA, with GRPO at $\leq$1200 steps, learning rates
$10^{-6}$--$10^{-4}$ and group sizes up to 50. Full-parameter training and prolonged-RL regimes \citep{liu2025prorl} remain open. The larger-rate matched cells are single runs. The stabilizers a practitioner would add to GRPO at such rates (a larger KL
weight, clipping, a length penalty, warm-up) were not run, and a nominal
learning rate is not the same step size for the two losses, so we report the remaining gap without attributing it to a cause. Two further choices are untested: the KL term of the
$\beta{=}0.01$ cells anchors to the base model, not to the start
checkpoint (TRL's reference for a LoRA policy), and matching steps and
prompts gives GRPO sixteen sampled completions per prompt where matched
SFT sees one written answer, so matching by oracle calls or tokens would be another defensible budget. Both domains are linear, steady-state systems
with synthetic text inputs, and heat conduction mirrors frames
structurally: the replication evidences recipe robustness, not domain
generality, and transfer to nonlinear or open-ended tasks is untested.

\textbf{Behavioral claims only.} ``Forward model'' is a behavioral
description (\S\ref{sec:task}). Linear probes do not adjudicate here
(\S\ref{sec:task}: they decode equally from every arm, including the
behaviorally inert \arm{O}, at the surface baseline)
\citep{hewitt2019control}. We make no mechanistic claim, and probing during generation is an open direction.

\textbf{Evaluation scope.} Post-edit structures belong to a family seen
in training (the out-of-format evidence is the action interface). The
readout matrix (\S\ref{sec:ladder}, Table~\ref{tab:readout}) scores
each arm on its own target as well as on ordinal questions. It is computed on seed-42 checkpoints and on the tier-1 streams, and every arm
predicts magnitudes less well than it orders them, so the value readouts
speak to ordering, not precision.

\textbf{One template and one in-context control.} Descriptions instantiate
a single fixed template---the price of byte-identical control, although results survive a paraphrased evaluation surface (Table~\ref{tab:d8}). The
frontier comparison is one
model, one $k$, text-only. Many-shot \citep{agarwal2024manyshot} and
tool use are untested---a tool-augmented model would likely solve the
task, but what the tool knows is not what the weights carry.

\textbf{Statistical scope.} All seven supervision arms carry three
training seeds under the calibrated protocol (Table~\ref{tab:d1}), as do
the dense-SFT, best-practice and lower-start GRPO endpoints of
\S\ref{sec:verifier}. The matched-recipe cells are single-seed, and the
contrasts of Table~\ref{tab:d7} compare single checkpoints on fresh
streams and are checkpoint-level evidence. ``Statistically
indistinguishable'' reports a failure to reject except where a
TOST-style bound is given (Table~\ref{tab:d7}).

\section*{Ethics Statement}
OraclePhys is fully synthetic: all structures are procedurally generated, and no human subjects, annotators, or personal data are involved
anywhere. The testbed and any released checkpoints are research
artifacts for studying training objectives and \textbf{must not be used
for real structural design, assessment, or code compliance}: the FE
oracle implements a simplified 2-D linear model that does not represent
engineering practice. One hazard is specific to the data release: the
OraclePhys-30K \arm{S} partition carries \emph{deliberately shuffled}
(false) physics labels---it exists as a placebo control and must never
be used as training supervision outside that role. The release builder
writes it only on request, into a separate directory beside a
do-not-train marker, and the datasheet lists it under controls. The release ships under a research license with
this intended-use statement attached, alongside the licenses of the
assets it builds on (OpenSees \citep{mckenna2011opensees}, Qwen3 \citep{yang2025qwen3}, Llama-3.1 \citep{grattafiori2024llama3}, subject to their
respective terms).

\bibliography{references}

\appendix
\section{Reproducibility}
All instance streams are deterministic, so every run answers the same questions. Every results file stores per-instance correctness arrays, and every number in the paper regenerates from the archived result files with a
single script, with all comparisons run as exact McNemar tests
\citep{mcnemar1947} with paired bootstrap CIs. Under greedy decoding,
re-running a bench reproduces aggregates exactly. Code, result files,
figure scripts, and the OraclePhys-30K supervision suite will be
released, together with the evaluation streams
exported as a static, FE-free benchmark (Appendix~\ref{app:testbed}):
prompt text plus oracle truths per instance, surface-formula reference rows
recomputed on the exported streams, and a SHA-256 over all truths as
integrity fingerprints---external models can be scored, and paired against
every model reported here, without installing the FE stack.

\begin{figure*}[t]
\centering
\begin{minipage}[t]{0.48\linewidth}
\centering
\includegraphics[width=0.9\linewidth]{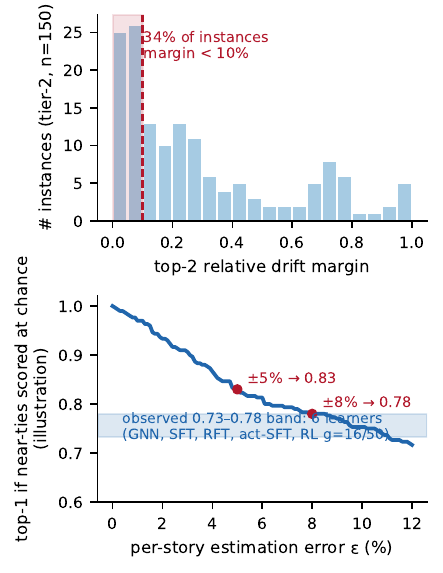}
\caption{Sensitivity illustration (\S\ref{sec:ceilings}). Top: 34\% of
tier-2 instances have a top-2 relative drift margin under 10\%. Bottom:
the top-1 accuracy that a hypothetical per-story estimation error
$\varepsilon$ would allow if every instance with margin below
$2\varepsilon$ were scored at chance. This is an illustration of the
margin structure, not an upper bound; the observed 0.73--0.78 band is
reported as an observation.}
\label{fig:wall}
\end{minipage}\hfill
\begin{minipage}[t]{0.48\linewidth}
\centering
\includegraphics[width=\textwidth]{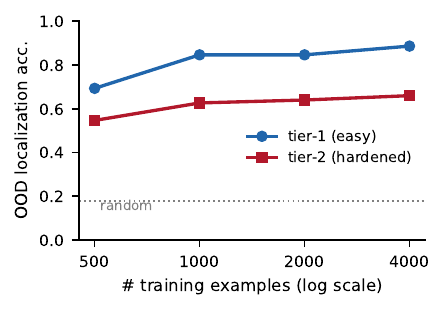}
\caption{Data efficiency: tier-1 saturates by $n{=}1$k; tier-2 gains
$+0.02$ at the last doubling, while one difficulty-matched second round
buys $+0.09$.}
\label{fig:ndata}
\end{minipage}
\end{figure*}

\begin{figure}[t]
\centering
\includegraphics[width=\columnwidth]{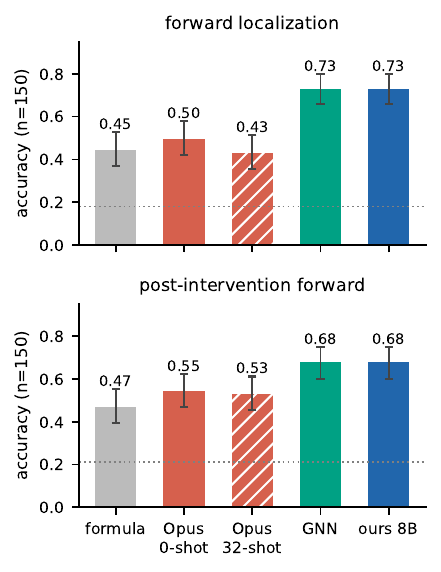}
\caption{Tier-2 comparison (\S\ref{sec:ceilings}), same 150 instances per
axis. 32 oracle-labeled in-context examples do not move Opus 4.8
($0.50 \to 0.43$); the trained 8B leads its best condition by $+0.23$
and its 32-shot condition by $+0.30$ ($p<10^{-4}$), with no detectable
difference from the specialist GNN. Error bars: 95\% Wilson intervals;
pairwise judgments: exact McNemar (Table~\ref{tab:d4}).}
\label{fig:frontier}
\end{figure}

\begin{figure}[t]
\centering
\includegraphics[width=\columnwidth]{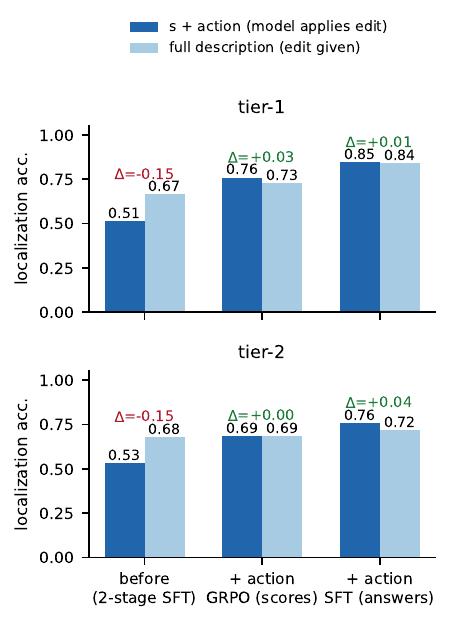}
\caption{Closing the action-application gap (\S\ref{sec:routing}). Left
group of each panel: the dense-SFT model predicts the post-edit response
from a description (light bar) but not from an action (dark bar; gap
$-0.15$). Action-format training closes the gap whether the oracle
scores the answers (GRPO, middle) or writes them (SFT, right)---and the
dense variant also raises the surrounding held-out accuracy.}
\label{fig:interface}
\end{figure}

\section{Testbed details}\label{app:testbed}

\textbf{Population and splits.} Problems specify geometry (bays, stories,
spans, story heights), constraints (forbidden brace cells), and the lateral
load (total base shear; vertical profile $F_x(s) \propto s^{\alpha}$).
Training problems have 3--4 stories; the OOD pool has 5--6 stories under
held-out geometry. Designs assign per-story column/beam sections from a
discrete catalog, a connection type (pinned or moment), braced cells, and
damage events.

\textbf{Tier-1 sampler.} Follows common design practice: near-monotone
column profiles, code-style bracing layouts, isolated damage---regularities
a surface reader can exploit (bag-of-words classifies the governing story at
0.63; geometry-only features 0.39; chance 0.18).

\textbf{Tier-2 hardening.} Three independent randomizations compose:
(i)~\emph{jittered column profiles}---a bounded random walk over the section
catalog replaces monotone profiles; (ii)~\emph{composed weakening
events}---\texttt{weak}, \texttt{twin}, \texttt{scatter}, \texttt{cutoff}
stack by independent coin flips (zero to four per structure), placing
braces and damage so that no single event determines the answer;
(iii)~\emph{load-shape pool}---$\alpha \in \{0.7, 1.0, 1.4, 2.0\}$, written
into the prompt. Verification: the bag-of-words probe falls to 0.26
(chance 0.18), geometry-only to 0.38, and the structured-feature probe from
0.76 to 0.57; ranking-target variants fall equally (rank-top-1
$0.57 \to 0.33$).

\textbf{The first-order formula.} Story shear $V_s = \sum_{k \geq s} F_k$
divided by a story-stiffness proxy aggregated from column sections (with a
connection-type factor), brace areas, and damage multipliers; stories are
ranked by $V_s / K_s$. It reads numeric section and brace properties from
the design object directly---information strictly more direct than the
prompt text---and scores 0.53/0.47 (tier-1) and 0.45/0.47 (tier-2)
localization/post-intervention.

\textbf{Heat-conduction isomorph.} 2-D steady-state conduction on a
grid with per-row materials from a conductivity catalog, high-conductivity
cooling cells, near-insulating defects, and configurable heat-sink
boundaries; the target is the per-row peak-temperature ranking. The mapping
is: rows $\leftrightarrow$ stories; material catalog $\leftrightarrow$
section catalog; cooling cells $\leftrightarrow$ braces; insulating defects
$\leftrightarrow$ damage; sink configuration $\leftrightarrow$ boundary
conditions; source-shape exponent $\leftrightarrow$ load shape. Its audit:
the 1-D two-sink ladder formula reaches 0.53 localization but only 0.21
post-intervention (2-D rerouting is invisible to it); prior 0.25; feature
MLP 0.46.

\textbf{Released benchmark.} The four description axes of both tiers are
exported as self-contained JSON (\texttt{frames\_tier\{1,2\}.json}), with
the heat-domain export, the 2-shot calibration prompts, and per-tier
fingerprints shipped alongside: 150
instances per axis (80 for extrapolation), each carrying the exact prompt
text and the oracle truths (governing story; full ranking where defined), so
external models are scored with a pure-\texttt{numpy} parser and no FE
dependency. Every instance asks the same ranking question: the readout
instrument is held fixed while the training-side objective varies; the
value and compliance readouts of Table~\ref{tab:readout} ship as a
separate stream set with per-instance arrays. Two integrity anchors ship
inside each file: the
surface-formula reference row recomputed on the exported stream must match
the paper's fingerprints (0.5333/0.4733 on tier~1), and a SHA-256 over all
truths pins the instance set. The invariance and action streams, which
require paired or counterfactual FE calls, regenerate deterministically from
the released code. Because per-instance correctness arrays for every model
reported here are also released, an external model can be compared against
any row of this paper by exact McNemar test without re-running our models.

\section{Training details}\label{app:training}

\textbf{Supervised arms.} Qwen3-8B \citep{yang2025qwen3} (and Llama-3.1-8B-Instruct \citep{grattafiori2024llama3} for the family
replication) with LoRA \citep{hu2022lora} $r{=}32$, $\alpha{=}64$, dropout 0.05; lr $10^{-4}$;
3 epochs; cutoff 2048; batch 2 $\times$ 4 accumulation; seeds 42, 43 and
44 for all seven frames arms. The three gate arms share byte-identical
structure text (asserted at build time with a token-level leakage audit) and
differ only in the question--answer suffix. The round-2 dense stage resumes
the tier-1 \arm{P} adapter on hardened-pool labels; the action-SFT control
resumes the round-2 adapter on 4k oracle-labeled (state, action)
$\to$ ranking pairs drawn from the same distribution as the action-GRPO
pool (uniform random story; connection-preserving brace edit).

\textbf{GRPO.} TRL \citep{vonwerra2020trl} implementation with the Dr.~GRPO loss
\citep{liu2025understanding}
(\texttt{loss\_type=dr\_grpo}, no length normalization) and
\texttt{scale\_rewards=False} (advantages not divided by group std);
temperature 1.0; lr $10^{-6}$ unless stated; completion cap 512 tokens
(128 in the $g{=}50$ cells, 768 with thinking); per-device batch 16 with
2-step accumulation (10 with 2- or 5-step accumulation in the $g{=}50$
cells, 1 with 8-step accumulation with thinking; Table~\ref{tab:grpo})
on RTX 3090s. Per-run settings are in
Table~\ref{tab:grpo}; the round-2, action-format and matched runs carried a
small KL anchor, $\beta{=}0.01$ (with LoRA, TRL's reference policy for
the KL term is the adapter-disabled base model, not the start
checkpoint), the rest none. Reward: $0.6 \cdot
(\rho_{\text{Spearman}}+1)/2 + 0.4 \cdot \mathbf{1}[\text{top-1 correct}]$,
with unparseable completions floored at $-0.5$. Budgets per protocol: the
dense round trains ${\sim}940$ optimizer steps at lr $10^{-4}$ (2 epochs
over 3{,}760 rows, batch $2\times4$); GRPO trains 400--1200 steps at lr
$10^{-6}$ ($10^{-5}$ in the lr probe; $10^{-5}$, $3{\times}10^{-5}$ and
$10^{-4}$ in the matched cells); the think-RL/nothink-RL contrast matches Y's steps and
configuration to X exactly. At lr $\geq 3{\times}10^{-5}$ the policy
occasionally emits a completion that runs to the 512-token cap, and the
logits of such a 16-sequence micro-batch exceed a 24 GB card. The
$3{\times}10^{-5}$ cell therefore resumed from its step-500 checkpoint
(optimizer, scheduler and random-number state restored) at per-device
batch 4 with 8-step accumulation for its last 100 steps, and the
$10^{-4}$ cell ran at that micro-batching throughout; the same 32
sequences enter every optimizer step, and since the Dr.~GRPO loss
divides the summed token loss by a fixed constant, the loss and gradient
of each step are unchanged.
 Total compute across
all training runs, seeds, and evaluations in the paper is on the order of
300 RTX-3090 GPU-hours; frontier-model calls total ${\sim}600$ API
requests. Round-1 diagnostics
recorded the within-group reward-std collapse ($0.122 \to 0.049$) that
motivated the difficulty-filtered round-2. Variants used in
\S\ref{sec:verifier}--\ref{sec:routing}: difficulty filtering (train only on
prompts whose greedy answer scores below a threshold), the action-format
pool, and the best-practice cell ($g{=}50$, five-way data parallel,
filter threshold 0.9).

\textbf{Frontier protocol.} API models are queried one instance per call
with responses disk-cached. The frontier baseline is
\texttt{claude-opus-4-8} via the Anthropic Messages API with extended
thinking enabled and provider-default decoding, queried on 2026-07-06 to
2026-07-08; the 32-shot condition prepends worked
train-split examples with oracle answers under prompt caching; outputs are
parsed by the shared parser and a parse failure scores as a miss. The
verification row of \S\ref{sec:stats} is recomputed on the identical stream
post hoc.

\textbf{OraclePhys-30K composition.} Seven answer-form arms $\times$
3{,}760 byte-identical training rows (26{,}320), plus action-interface
supervision (${\sim}4$k pairs) and the hardened second-round curriculum
(\S\ref{sec:verifier}); every label except the
\arm{S} placebo is written programmatically from the oracle's solution.

\textbf{Arm and checkpoint glossary.} Training arms: \arm{P} ranking;
\arm{A} peak scalar; \arm{O} compliance boolean; \arm{K} pointwise scalar;
\arm{V} value vector; \arm{G} governing story only; \arm{S} shuffled
placebo. Checkpoints: \emph{SFT-hard} = \arm{P} + dense hardened round;
\emph{RFT} = \arm{P} + filtered self-distillation; \emph{Y} = SFT-hard +
no-think GRPO; \emph{X} = think-bridge + GRPO pipeline; \emph{BP-GRPO} =
SFT-hard + $g{=}50$ filtered GRPO; \emph{final} = SFT-hard + action-GRPO
(the action-GRPO assembly); \emph{action-SFT} = SFT-hard + action-label SFT
(the pure-supervised assembly, deployed flagship).

\textbf{An example instance (tier-1 forward stream, verbatim).}
\begin{quote}\small\ttfamily
A 3-bay, 5-story steel frame (26 ft bays, 13 ft story height) under lateral
seismic load (total base shear 74 kip, increasing with height).\\
- beam (all floors): W21x68\\
- columns by story: story 1: W12x53; story 2: W12x53; story 3: W8x24;
story 4: W12x53; story 5: W12x53\\
- beam-column connections: moment (rigid joints)\\
- braced cells: story 1: bays [1, 2, 3]; story 2: bays [1, 2, 3]; story 3:
bays [1, 3]; story 4: bays [1, 2, 3]\\[2pt]
Rank ALL 5 stories from LARGEST to SMALLEST inter-story drift ratio (story
1 = bottom, story 5 = top). Respond with JSON only: \{"ranking": [story
numbers, largest-drift first]\}.
\end{quote}
Oracle answer: ranking $[5,3,2,4,1]$, governing story 5: the only
unbraced story sits at the top of a height-weighted load, yet story 3's
section cut plus missing brace nearly overtake it---the ranking requires
weighing both effects, and no single sentence carries the answer. Descriptions are templated with
randomized structure; surface-cue exploitability is audited and hardened in
tier-2 (\S\ref{sec:tiers}).

\textbf{Thinking bridge.} 4k programmatic rationales ("evidence lines
$\to$ verdict $\to$ answer," generated from structure features and FE ground
truth; ${\sim}170$ tokens) fine-tuned on the dense-SFT model restore
think-mode termination (zero clipping from 100\%). X continues this
checkpoint with GRPO under thinking; Y applies matched-compute GRPO to the
same parent without thinking.

\section{Alternative-explanation sweep for \S4}\label{app:alt}

\textbf{Reward design.} The reward (Appendix~\ref{app:training}) is graded,
not binary: 60\% continuous full-profile rank correlation plus 40\% explicit
localization credit, with a parse floor that preserves gradient toward
well-formed answers. The two densifications a critic would propose---reward
the profile shape; reward localization directly---are both already present.
At a matched learning rate the scalar reward moved forward accuracy as
far as the label did (Table~\ref{tab:matched}; within 0.03 on the pooled
streams, Table~\ref{tab:d7}), so reward shape is not what limited the
standard-rate runs.

\begin{figure}[t]
\centering
\includegraphics[width=\columnwidth]{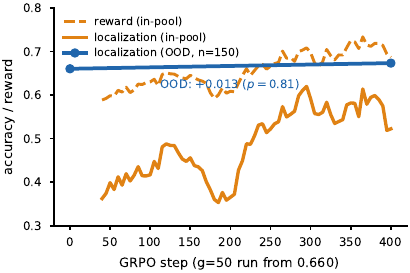}
\caption{In-pool gain without held-out gain at lr $10^{-6}$: over the
lower-start $g{=}50$ run, reward and in-pool localization climb while
OOD localization stands still
($+0.013$, $p{=}0.81$). Curves: rolling mean over 8 logged steps.}
\label{fig:judge}
\end{figure}

\textbf{Optimizer health.} Over GRPO training the reward climbs
$0.80 \to 0.93$ and in-pool localization $0.74 \to 0.88$ (over 50
samples per hardened prompt from the start checkpoint, pass@1 is 0.796
and pass@50 is 0.941) while OOD
localization moves $+0.01$: optimization succeeds at its stated objective,
and at this learning rate the objective is satisfiable without
transferable physics; at a tenfold rate from the \arm{P}-arm start (the matched cell) the same objective transfers,
and at $3{\times}10^{-5}$ and $10^{-4}$ the policy loses held-out
accuracy while emitting runaway completions (Table~\ref{tab:matched};
Appendix~\ref{app:training}). The
lower-start relaunch reproduces the pattern:
reward $0.62 \to 0.69$, in-pool localization $0.41 \to 0.55$ (first- vs.\
last-40-step means), OOD $+0.013$.

\textbf{Exploration.} The round-2 recipe trains exclusively on unsolved
prompts (greedy-score filtering) with $g{=}16$ samples at temperature 1.0;
OOD localization reaches 0.687 from the 0.660 start (round-1: 0.667;
$p{=}0.15$ vs.\ the dense control's 0.747).

\textbf{Capacity.} The identical LoRA configuration absorbs $+0.09$ to
$+0.14$ on every axis when the oracle writes answers
(Table~\ref{tab:twouses}), so parametric capacity does not bind.

\textbf{Learning rate.} The matched cells (Table~\ref{tab:matched}) are
the reading that accounts for most of the pattern: at $10^{-6}$ neither
update moves held-out accuracy, at $10^{-5}$ both reach forward 0.760
from the same start and stay within 0.03 on the pooled streams, and at
$3{\times}10^{-5}$ and $10^{-4}$ GRPO falls below its start (forward 0.627
and 0.487 against 0.660) while SFT from the same start and budget
reaches 0.767 and 0.687 and stays above the start on the pooled streams
(Table~\ref{tab:d7}); dense SFT at $10^{-4}$ with a bigger batch and a
second epoch reaches 0.747. Whether a differently stabilized GRPO could
use those rates is untested.

\section{Thinking-channel details (\S\ref{sec:routing})}\label{app:think}
Direct-answer SFT leaves the model unable to use its thinking mode: asked to
reason before answering, 100\% of generations hit the token limit without
closing the think block. The rationale bridge is 4k automatically generated
chains of the form ``evidence lines $\to$ verdict $\to$ answer''---evidence
lines read off structure features (sections, bracing, damage, load shape),
the verdict names the governing story, the answer repeats the JSON
ranking---fine-tuned on top of the dense-SFT checkpoint with the same LoRA
configuration. After the bridge, thinking terminates in ${\sim}170$ tokens
with zero clipping, and length stays stable (${\sim}172$ tokens) throughout
subsequent RL, so the on/off contrast in \S\ref{sec:think} runs on identical
weights with a functioning channel in both conditions.

\textbf{The three \S\ref{sec:think} contrasts} ($n{=}150$, exact McNemar):
(i)~bridge cost: thinking off, the bridged model scores 0.673 vs.\ its
0.747 parent ($-0.073$, discordant 3:14, $p{=}0.013$); (ii)~thinking on:
forward $+0.033$ ($p{=}0.33$), post-intervention $-0.013$ ($p{=}0.81$);
(iii)~think-RL vs.\ nothink-RL from the same SFT-hard ancestor (\arm{X}
via the bridge, \arm{Y} directly; the gap is commensurate with the bridge
tax of (i)): 0.707 vs.\ 0.760,
95\% CI $[-0.133,+0.027]$, $p{=}0.27$, the point estimate favoring not
thinking. RL preserves the channel (thinking length stable throughout); it
never pays for itself.

\section{Full results tables}\label{app:tables}

Tables~\ref{tab:d1}--\ref{tab:d8} report the core per-arm results across
axes and tiers, and the key paired comparisons with discordant counts.
Table~\ref{tab:readout} reports the training-target by readout matrix,
and Table~\ref{tab:cases} four held-out cases with every arm's verbatim
output. Every value regenerates from the archived result files via
\texttt{paper/gen\_appendix.py} and \texttt{paper/gen\_readout\_table.py}.

\begin{table*}[t]\centering
\caption{Frames, tier-1 (n=150 per axis, 80 for extrapolation; localization unless noted). Gate
arms are 3-seed mean$\pm$sample sd; other raw rows single runs on the same stream.
K = pointwise-scalar control; GRPO-from-scratch was evaluated on this
tier's stream (tier-2 forward 0.167, text only). Raw rows (top) conflate missing knowledge with output-format
failure (\S3.3); the fmt2 block repeats every arm under the shared 2-shot
format calibration, every supervision arm as a 3-seed mean$\pm$sample sd
(seeds 42--44)---cross-arm judgments use these rows (Table~\ref{tab:d4}).}
\label{tab:d1}\vspace{1.2ex}\small
\begin{tabular}{lcccccc}
\toprule
model & forward & $\rho$ & extrap & two-step & post-int. & invar. \\
\midrule
\arm{P} (3 seeds) & 0.86$\pm$0.02 & 0.90$\pm$0.00 & 0.82$\pm$0.02 & 0.72$\pm$0.01 & 0.71$\pm$0.03 & 1.00$\pm$0.00 \\
\arm{A} (3 seeds) & 0.56$\pm$0.06 & 0.58$\pm$0.08 & 0.61$\pm$0.09 & 0.35$\pm$0.09 & 0.29$\pm$0.05 & 0.99$\pm$0.01 \\
\arm{O} (3 seeds) & 0.15$\pm$0.07 & -0.02$\pm$0.22 & 0.17$\pm$0.06 & 0.21$\pm$0.00 & 0.12$\pm$0.00 & 0.60$\pm$0.24 \\
\arm{K} (pointwise scalar) & 0.56 & 0.64 & 0.61 & 0.45 & 0.35 & 1.00 \\
\arm{V} (value vector) & 0.52 & 0.52 & 0.55 & 0.25 & 0.30 & 0.87 \\
\arm{G} (governing only) & 0.85 & 0.84 & 0.82 & 0.74 & 0.70 & 1.00 \\
\arm{S} (shuffled placebo) & 0.24 & -0.07 & 0.20 & 0.15 & 0.04 & 0.49 \\
untrained base & 0.25 & -0.09 & 0.31 & 0.13 & 0.08 & 0.29 \\
final model & 0.84 & 0.91 & 0.80 & 0.67 & 0.73 & 1.00 \\
action-SFT & 0.85 & 0.91 & 0.84 & 0.79 & 0.87 & 1.00 \\
GRPO from scratch & 0.43 & 0.17 & 0.41 & 0.14 & 0.09 & 0.53 \\
\midrule
fmt2: \arm{P} (3 seeds) & 0.74$\pm$0.01 & 0.85$\pm$0.02 & 0.78$\pm$0.01 & 0.62$\pm$0.04 & 0.74$\pm$0.04 & 1.00$\pm$0.00 \\
fmt2: \arm{A} (3 seeds) & 0.50$\pm$0.09 & 0.69$\pm$0.04 & 0.53$\pm$0.07 & 0.42$\pm$0.02 & 0.47$\pm$0.04 & 1.00$\pm$0.00 \\
fmt2: \arm{O} (3 seeds) & 0.32$\pm$0.05 & 0.26$\pm$0.01 & 0.31$\pm$0.04 & 0.28$\pm$0.05 & 0.18$\pm$0.01 & 0.88$\pm$0.18 \\
fmt2: \arm{K} (3 seeds) & 0.56$\pm$0.07 & 0.70$\pm$0.01 & 0.59$\pm$0.02 & 0.35$\pm$0.02 & 0.36$\pm$0.01 & 1.00$\pm$0.00 \\
fmt2: \arm{V} (3 seeds) & 0.64$\pm$0.03 & 0.72$\pm$0.02 & 0.70$\pm$0.05 & 0.41$\pm$0.03 & 0.39$\pm$0.08 & 1.00$\pm$0.00 \\
fmt2: \arm{G} (3 seeds) & 0.75$\pm$0.07 & 0.77$\pm$0.04 & 0.75$\pm$0.04 & 0.67$\pm$0.05 & 0.69$\pm$0.02 & 1.00$\pm$0.00 \\
fmt2: \arm{S} (3 seeds) & 0.37$\pm$0.03 & 0.22$\pm$0.02 & 0.41$\pm$0.01 & 0.19$\pm$0.06 & 0.11$\pm$0.01 & 0.57$\pm$0.04 \\
fmt2: untrained base & 0.30 & 0.25 & 0.29 & 0.28 & 0.16 & 0.77 \\
\midrule
Llama-3.1 \arm{P} & 0.91 & 0.90 & 0.88 & 0.78 & 0.73 & 1.00 \\
Llama-3.1 \arm{A} & 0.11 & -0.24 & 0.09 & 0.17 & 0.12 & 0.41 \\
Llama-3.1 base & 0.19 & -0.11 & 0.21 & 0.09 & 0.23 & 0.40 \\
fmt2: Llama-3.1 \arm{P} & 0.84 & 0.88 & 0.78 & 0.69 & 0.68 & 1.00 \\
fmt2: Llama-3.1 \arm{A} & 0.25 & 0.25 & 0.15 & 0.35 & 0.39 & 0.60 \\
fmt2: Llama-3.1 base & 0.19 & -0.28 & 0.23 & 0.15 & 0.29 & 0.62 \\
\midrule
first-order formula & 0.53 & 0.64 & 0.62 & 0.53 & 0.47 & 1.00 \\
GNN (graph input) & 0.82 & 0.93 & 0.82 & 0.67 & 0.63 & 1.00 \\
feature MLP & 0.19 & 0.47 & 0.21 & 0.21 & 0.27 & 1.00 \\
Opus 4.8, 0-shot & 0.63 & 0.69 & -- & -- & 0.50 & -- \\
\bottomrule\end{tabular}\end{table*}
\begin{table*}[t]\centering
\caption{Frames, tier-2 (hardened; n=150 per axis, 80 for extrapolation).}
\label{tab:d2}\vspace{1.2ex}\small
\begin{tabular}{lcccccc}
\toprule
model & forward & $\rho$ & extrap & two-step & post-int. & invar. \\
\midrule
\arm{P} (tier-1 weights) & 0.66 & 0.69 & 0.57 & 0.59 & 0.55 & 1.00 \\
+ GRPO round-1 & 0.67 & 0.66 & 0.57 & 0.57 & 0.57 & 1.00 \\
+ GRPO round-2 (filtered) & 0.69 & 0.69 & 0.59 & 0.59 & 0.63 & 1.00 \\
+ dense SFT (SFT-hard) & 0.75 & 0.78 & 0.71 & 0.71 & 0.69 & 1.00 \\
+ RFT (filter-then-imitate) & 0.75 & 0.78 & 0.69 & 0.67 & 0.68 & 1.00 \\
SFT-hard + nothink-RL & 0.76 & 0.79 & 0.64 & 0.64 & 0.65 & 1.00 \\
SFT-hard + best-practice RL (g=50; seed 42 of 3, mean .751) & 0.78 & 0.79 & 0.72 & 0.67 & 0.71 & 1.00 \\
\arm{P} + full-recipe RL from 0.660 (g=50) & 0.67 & 0.66 & 0.57 & 0.53 & 0.61 & 1.00 \\
final model (action-format GRPO) & 0.73 & 0.77 & 0.64 & 0.65 & 0.68 & 1.00 \\
action-SFT & 0.74 & 0.74 & 0.69 & 0.71 & 0.71 & 1.00 \\
think-RL pipeline & 0.71 & 0.74 & 0.68 & 0.63 & 0.68 & 1.00 \\
\midrule
first-order formula & 0.45 & 0.51 & 0.36 & 0.55 & 0.47 & 1.00 \\
GNN (graph input) & 0.73 & 0.82 & 0.68 & 0.73 & 0.68 & 1.00 \\
Opus 4.8, 0-shot & 0.50 & 0.59 & -- & -- & 0.55 & -- \\
Opus 4.8, 32-shot & 0.43 & 0.32 & -- & -- & 0.53 & -- \\
\bottomrule\end{tabular}\end{table*}
\begin{table*}[t]\centering
\caption{Heat-conduction isomorph (n=150 per axis, 80 for extrapolation; single seed).
The frames-trained \arm{P} arm transferred zero-shot to heat localizes at 0.133
(uniform guess 0.187) while its output format survives intact.}
\label{tab:d3}\vspace{1.2ex}\small
\begin{tabular}{lcccccc}
\toprule
model & forward & $\rho$ & extrap & two-step & post-int. & invar. \\
\midrule
\arm{P} & 0.89 & 0.93 & 0.79 & 0.82 & 0.73 & 1.00 \\
\arm{A} & 0.29 & -0.11 & 0.21 & 0.23 & 0.15 & 0.93 \\
\arm{O} & 0.33 & -0.15 & 0.25 & 0.24 & 0.11 & 0.95 \\
untrained base & 0.25 & -0.07 & 0.30 & 0.30 & 0.14 & 0.69 \\
\midrule
1-D ladder formula & 0.53 & 0.79 & 0.65 & 0.54 & 0.21 & 1.00 \\
feature MLP & 0.46 & 0.57 & 0.29 & 0.37 & 0.49 & 0.95 \\
\bottomrule\end{tabular}\end{table*}
\begin{table*}[t]\centering
\caption{Key paired comparisons (tier-2 unless noted; rows prefixed
``fmt2:'' are the \emph{tier-1} stream under the 2-shot format calibration
of \S3.3; exact McNemar on the shared deterministic streams; 95\%
paired-bootstrap CI on the difference; $^*$: $p<0.05$; $q$:
Benjamini--Hochberg FDR across all 32 tests in this
table).}
\label{tab:d4}\vspace{1.2ex}\footnotesize
\begin{tabular}{lcccccc}
\toprule
comparison & A & B & $\Delta$ [CI] & discord & $p$ & $q$ \\
\midrule
final vs.\ Opus 0-shot (fwd) & 0.733 & 0.500 & +0.233 [+0.15,+0.32] & 44:9 & 0.0000$^*$ & 0.000 \\
final vs.\ Opus 32-shot (fwd) & 0.733 & 0.433 & +0.300 [+0.21,+0.39] & 53:8 & 0.0000$^*$ & 0.000 \\
Opus 32- vs.\ 0-shot (fwd) & 0.433 & 0.500 & -0.067 [-0.13,+0.00] & 9:19 & 0.0872 & 0.179 \\
Opus 0-shot vs.\ formula (fwd) & 0.500 & 0.447 & +0.053 [+0.00,+0.11] & 13:5 & 0.0963 & 0.181 \\
final vs.\ GNN (fwd) & 0.733 & 0.733 & +0.000 [-0.07,+0.07] & 16:16 & 1.0000 & 1.000 \\
final vs.\ GNN (post-int.) & 0.680 & 0.680 & +0.000 [-0.09,+0.09] & 22:22 & 1.0000 & 1.000 \\
fmt2: \arm{P} vs.\ \arm{A} (fwd) & 0.733 & 0.467 & +0.267 [+0.16,+0.37] & 60:20 & 0.0000$^*$ & 0.000 \\
fmt2: \arm{K} vs.\ \arm{A} (fwd) & 0.607 & 0.467 & +0.140 [+0.07,+0.21] & 25:4 & 0.0001$^*$ & 0.001 \\
fmt2: \arm{P} vs.\ \arm{K} (fwd) & 0.733 & 0.607 & +0.127 [+0.03,+0.22] & 37:18 & 0.0145$^*$ & 0.042 \\
fmt2: \arm{K} vs.\ \arm{A} (post-int.) & 0.353 & 0.420 & -0.067 [-0.13,-0.01] & 5:15 & 0.0414$^*$ & 0.102 \\
fmt2: \arm{A} vs.\ base (post-int.) & 0.420 & 0.160 & +0.260 [+0.19,+0.33] & 42:3 & 0.0000$^*$ & 0.000 \\
fmt2: \arm{O} vs.\ base (fwd) & 0.327 & 0.300 & +0.027 [-0.01,+0.07] & 6:2 & 0.2891 & 0.402 \\
fmt2: \arm{V} vs.\ \arm{P} (fwd) & 0.613 & 0.733 & -0.120 [-0.21,-0.03] & 14:32 & 0.0114$^*$ & 0.040 \\
fmt2: \arm{V} vs.\ \arm{K} (fwd) & 0.613 & 0.607 & +0.007 [-0.06,+0.07] & 13:12 & 1.0000 & 1.000 \\
fmt2: \arm{G} vs.\ \arm{P} (fwd) & 0.680 & 0.733 & -0.053 [-0.14,+0.03] & 16:24 & 0.2682 & 0.390 \\
fmt2: \arm{S} vs.\ base (fwd) & 0.353 & 0.300 & +0.053 [-0.01,+0.12] & 16:8 & 0.1516 & 0.255 \\
fmt2: \arm{V} vs.\ \arm{A} (post-int.) & 0.313 & 0.420 & -0.107 [-0.19,-0.03] & 10:26 & 0.0113$^*$ & 0.040 \\
RFT vs.\ its start (fwd) & 0.753 & 0.660 & +0.093 [+0.03,+0.15] & 18:4 & 0.0043$^*$ & 0.020 \\
RFT vs.\ SFT-hard (fwd) & 0.753 & 0.747 & +0.007 [-0.04,+0.05] & 7:6 & 1.0000 & 1.000 \\
RFT vs.\ GRPO round-2, same start (fwd) & 0.753 & 0.687 & +0.067 [+0.01,+0.13] & 17:7 & 0.0639 & 0.146 \\
GRPO round-2 vs.\ start (post-int.) & 0.633 & 0.553 & +0.080 [+0.02,+0.15] & 18:6 & 0.0227$^*$ & 0.060 \\
BP-GRPO (g=50) vs.\ its start (fwd) & 0.780 & 0.747 & +0.033 [-0.01,+0.07] & 7:2 & 0.1797 & 0.287 \\
BP-GRPO vs.\ nothink-RL (g=50 vs.\ g=16) & 0.780 & 0.760 & +0.020 [-0.03,+0.07] & 9:6 & 0.6072 & 0.810 \\
nothink-RL vs.\ SFT-hard (fwd) & 0.760 & 0.747 & +0.013 [-0.04,+0.07] & 10:8 & 0.8145 & 0.965 \\
lower-start GRPO (g=50, from 0.660) vs.\ start (fwd) & 0.673 & 0.660 & +0.013 [-0.04,+0.07] & 10:8 & 0.8145 & 0.965 \\
lower-start GRPO vs.\ dense SFT, same start (fwd) & 0.673 & 0.747 & -0.073 [-0.15,+0.01] & 15:26 & 0.1173 & 0.208 \\
lower-start GRPO vs.\ GRPO round-2, same start (fwd) & 0.673 & 0.687 & -0.013 [-0.05,+0.03] & 4:6 & 0.7539 & 0.965 \\
bridge tax: think-SFT(off) vs.\ SFT-hard & 0.673 & 0.747 & -0.073 [-0.13,-0.02] & 3:14 & 0.0127$^*$ & 0.041 \\
think-RL vs.\ nothink-RL (fwd) & 0.707 & 0.760 & -0.053 [-0.13,+0.03] & 16:24 & 0.2682 & 0.390 \\
actSFT vs.\ SFT-hard act-axis & 0.760 & 0.533 & +0.227 [+0.15,+0.31] & 41:7 & 0.0000$^*$ & 0.000 \\
actSFT vs.\ final act-axis & 0.760 & 0.687 & +0.073 [-0.01,+0.15] & 23:12 & 0.0895 & 0.179 \\
final act vs.\ desc (t2) & 0.687 & 0.687 & +0.000 [-0.07,+0.07] & 15:15 & 1.0000 & 1.000 \\
\bottomrule\end{tabular}\end{table*}
\begin{table*}[t]\centering
\caption{Supplementary paired rows, added after the Table~\ref{tab:d4}
family was fixed; reported with raw $p$ only, outside that table's FDR
correction.}
\label{tab:d5}\vspace{1.2ex}\footnotesize
\begin{tabular}{lccccc}
\toprule
comparison & A & B & $\Delta$ [CI] & discord & $p$ \\
\midrule
fmt2: Llama \arm{P} vs.\ \arm{A} (fwd) & 0.840 & 0.247 & +0.593 [+0.51,+0.67] & 92:3 & 0.0000 \\
BP-GRPO vs.\ SFT-hard (two-step) & 0.673 & 0.707 & -0.033 [-0.08,+0.01] & 4:9 & 0.2668 \\
lower-start GRPO vs.\ its start (two-step) & 0.527 & 0.593 & -0.067 [-0.14,+0.01] & 12:22 & 0.1214 \\
Opus 32-shot vs.\ formula (fwd) & 0.433 & 0.447 & -0.013 [-0.09,+0.06] & 14:16 & 0.8555 \\
Opus 0-shot vs.\ formula (tier-1 fwd) & 0.633 & 0.533 & +0.100 [+0.05,+0.15] & 15:0 & 0.0001 \\
Opus 0-shot vs.\ formula (post-int.) & 0.547 & 0.473 & +0.073 [+0.02,+0.13] & 14:3 & 0.0127 \\
final vs.\ GNN (tier-1 post-int.) & 0.727 & 0.627 & +0.100 [+0.01,+0.19] & 35:20 & 0.0581 \\
action-SFT vs.\ final (tier-1 post-int.) & 0.867 & 0.727 & +0.140 [+0.07,+0.21] & 26:5 & 0.0002 \\
fmt2: \arm{P} vs.\ \arm{A} (post-int.) & 0.713 & 0.420 & +0.293 [+0.21,+0.38] & 50:6 & 0.0000 \\
fmt2: \arm{A} vs.\ base (fwd) & 0.467 & 0.300 & +0.167 [+0.11,+0.23] & 27:2 & 0.0000 \\
final vs.\ GNN (tier-1 fwd) & 0.840 & 0.820 & +0.020 [-0.05,+0.09] & 15:12 & 0.7011 \\
lr probe ($10^{-5}$, $g{=}50$, from dense SFT) vs.\ its start (4 axes pooled) & 0.642 & 0.715 & -0.074 [-0.12,-0.03] & 45:84 & 0.0008 \\
\bottomrule\end{tabular}\end{table*}
\begin{table*}[t]\centering
\caption{Action axis, tier-2 (act = one-sentence action applied internally;
desc = same transition given as a full post-edit description; paired tests
for these cells appear in Table~\ref{tab:d4}).}
\label{tab:d6}\vspace{1.2ex}\small
\begin{tabular}{lccc}
\toprule
model & act & desc & act$-$desc \\
\midrule
SFT-hard & 0.533 & 0.680 & -0.147 \\
action-SFT & 0.760 & 0.720 & +0.040 \\
final model & 0.687 & 0.687 & +0.000 \\
\bottomrule\end{tabular}\end{table*}
\begin{table*}[t]\centering
\caption{Powered direct endpoint contrasts on fresh $n{=}600$ tier-2
streams, three disjoint tier-2 streams pooled per comparison as
independent pairs (forward, two-step, post-intervention; extrapolation
excluded as a subset of the forward stream; exact McNemar; 95\%
paired-bootstrap CI; TOST-style equivalence = CI within $\pm0.05$; the
margin was fixed before the matched cells were run and is below the
smallest gain the paper interprets, dense SFT over its start).
Every contrast compares single seed-42 checkpoints and is
checkpoint-level evidence: instance-level pairing controls sampling
noise on the stream, not variation across training runs. Three training
seeds exist for the dense-SFT endpoint and for the best-practice and
lower-start GRPO endpoints (\S\ref{sec:verifier}); the RFT, GRPO
round-2 and matched-recipe SFT and GRPO endpoints and these pairings
themselves are single-run. Post-hoc
powered runs, reported outside the Table~\ref{tab:d4} FDR family.}
\label{tab:d7}\vspace{1.2ex}\scriptsize\setlength{\tabcolsep}{3.5pt}
\begin{tabular}{lccccccc}
\toprule
comparison & $n$ & A & B & $\Delta$ [CI] & discord & $p$ & equiv. \\
\midrule
dense SFT vs.\ GRPO round-2, same start & 1800 & 0.716 & 0.629 & +0.087 [+0.06,+0.11] & 287:131 & 1.7e-14 & no \\
dense SFT vs.\ lower-start GRPO, same start & 1800 & 0.716 & 0.611 & +0.105 [+0.08,+0.13] & 336:147 & 4.4e-18 & no \\
RFT vs.\ GRPO round-2, same start & 1800 & 0.710 & 0.629 & +0.081 [+0.06,+0.10] & 256:110 & 1.6e-14 & no \\
lower-start GRPO vs.\ its start & 1800 & 0.611 & 0.622 & -0.012 [-0.03,+0.01] & 155:176 & 0.27 & yes \\
RFT vs.\ dense SFT & 1800 & 0.710 & 0.716 & -0.006 [-0.02,+0.01] & 112:122 & 0.56 & yes \\
matched SFT ($10^{-5}$) vs.\ GRPO round-2 ($10^{-6}$), same start & 1800 & 0.666 & 0.629 & +0.037 [+0.02,+0.05] & 150:84 & 1.9e-05 & no \\
matched SFT ($10^{-5}$) vs.\ dense SFT & 1800 & 0.666 & 0.716 & -0.050 [-0.07,-0.03] & 99:189 & 1.2e-07 & no \\
matched SFT ($10^{-5}$) vs.\ its start & 1800 & 0.666 & 0.622 & +0.043 [+0.03,+0.06] & 156:78 & 3.8e-07 & no \\
matched SFT ($10^{-5}$) vs.\ matched GRPO ($10^{-5}$), same start & 1800 & 0.666 & 0.640 & +0.026 [+0.01,+0.05] & 196:150 & 0.015 & yes \\
matched GRPO ($10^{-5}$) vs.\ its start & 1800 & 0.640 & 0.622 & +0.018 [-0.01,+0.04] & 257:225 & 0.16 & yes \\
matched GRPO ($10^{-5}$) vs.\ dense SFT & 1800 & 0.640 & 0.716 & -0.076 [-0.10,-0.05] & 138:274 & 1.9e-11 & no \\
matched SFT ($10^{-5}$) vs.\ matched GRPO ($3{\times}10^{-5}$), same start & 1800 & 0.666 & 0.571 & +0.095 [+0.07,+0.12] & 325:154 & 4.1e-15 & no \\
matched GRPO ($3{\times}10^{-5}$) vs.\ its start & 1800 & 0.571 & 0.622 & -0.052 [-0.08,-0.03] & 252:345 & 0.00016 & no \\
matched GRPO ($3{\times}10^{-5}$) vs.\ dense SFT & 1800 & 0.571 & 0.716 & -0.145 [-0.17,-0.12] & 130:391 & 2.3e-31 & no \\
matched SFT ($10^{-5}$) vs.\ matched GRPO ($10^{-4}$), same start & 1800 & 0.666 & 0.434 & +0.231 [+0.21,+0.26] & 514:98 & 5.3e-69 & no \\
matched GRPO ($10^{-4}$) vs.\ its start & 1800 & 0.434 & 0.622 & -0.188 [-0.21,-0.16] & 118:456 & 7.9e-48 & no \\
matched GRPO ($10^{-4}$) vs.\ dense SFT & 1800 & 0.434 & 0.716 & -0.281 [-0.30,-0.26] & 54:560 & 4.6e-107 & no \\
matched SFT ($3{\times}10^{-5}$) vs.\ its start & 1800 & 0.672 & 0.622 & +0.049 [+0.03,+0.07] & 203:114 & 6.6e-07 & no \\
matched SFT ($3{\times}10^{-5}$) vs.\ matched GRPO ($3{\times}10^{-5}$), same start & 1800 & 0.672 & 0.571 & +0.101 [+0.08,+0.12] & 348:166 & 7.1e-16 & no \\
matched SFT ($3{\times}10^{-5}$) vs.\ dense SFT & 1800 & 0.672 & 0.716 & -0.044 [-0.06,-0.03] & 80:159 & 3.6e-07 & no \\
matched SFT ($10^{-4}$) vs.\ its start & 1800 & 0.662 & 0.622 & +0.040 [+0.02,+0.06] & 230:158 & 0.0003 & no \\
matched SFT ($10^{-4}$) vs.\ matched GRPO ($10^{-4}$), same start & 1800 & 0.662 & 0.434 & +0.228 [+0.20,+0.25] & 503:93 & 5.2e-69 & no \\
matched SFT ($10^{-4}$) vs.\ dense SFT & 1800 & 0.662 & 0.716 & -0.053 [-0.07,-0.03] & 108:204 & 6e-08 & no \\
matched SFT ($10^{-6}$) vs.\ GRPO round-2 ($10^{-6}$), same start & 1800 & 0.637 & 0.629 & +0.008 [-0.01,+0.02] & 116:101 & 0.34 & yes \\
matched SFT ($10^{-6}$) vs.\ its start & 1800 & 0.637 & 0.622 & +0.015 [+0.01,+0.02] & 51:24 & 0.0024 & yes \\
GRPO round-2 vs.\ its start (post-int.\ only) & 600 & 0.612 & 0.595 & +0.017 [-0.01,+0.05] & 51:41 & 0.35 & yes \\
final model vs.\ GNN (fwd$+$post-int.) & 1200 & 0.703 & 0.677 & +0.025 [-0.00,+0.06] & 186:156 & 0.12 & no \\
\bottomrule\end{tabular}\end{table*}
\begin{table*}[t]\centering
\caption{Paraphrased-template evaluation (tier-1 unless noted): the same
held-out instances re-rendered through a second linguistic surface
(running prose, reordered fields, reworded connection/bracing/load
phrasings; identical information, question, and parser). Frozen
checkpoints; no re-training.}
\label{tab:d8}\vspace{1.2ex}\small
\begin{tabular}{lcccccc}
\toprule
model & forward & $\rho$ & extrap & two-step & post-int. & invar. \\
\midrule
\arm{P} (para) & 0.85 & 0.89 & 0.82 & 0.63 & 0.69 & 1.00 \\
\arm{A} (para) & 0.35 & 0.30 & 0.36 & 0.23 & 0.12 & 0.99 \\
\arm{O} (para) & 0.22 & 0.10 & 0.20 & 0.21 & 0.13 & 0.90 \\
untrained base (para) & 0.33 & 0.16 & 0.31 & 0.14 & 0.10 & 0.59 \\
fmt2: \arm{P} (para) & 0.75 & 0.85 & 0.74 & 0.61 & 0.71 & 1.00 \\
fmt2: \arm{A} (para) & 0.44 & 0.62 & 0.44 & 0.43 & 0.37 & 1.00 \\
fmt2: \arm{O} (para) & 0.31 & 0.29 & 0.28 & 0.27 & 0.18 & 0.95 \\
fmt2: base (para) & 0.29 & 0.26 & 0.28 & 0.24 & 0.13 & 0.80 \\
\midrule
SFT-hard, tier-2 (para) & 0.69 & 0.76 & 0.64 & 0.66 & 0.61 & 0.99 \\
\bottomrule\end{tabular}\end{table*}

\begin{table*}[t]\centering
\caption{Training-target by readout matrix (\S\ref{sec:ladder}): every
supervision arm scored under every readout, on the four tier-1 evaluation
streams ($n{=}150$; 80 for extrapolation), seed-42 checkpoints. Each cell
reads \emph{with}/\emph{without} two worked examples (the paper's
calibrated protocol / none). Ordinal readouts report governing-story top-1
accuracy: \emph{rank} = full ranking, \emph{gov} = governing story only,
\emph{val-argmax} = argmax of the arm's own per-story values. Numerical
readouts report the median relative error of the predicted quantity:
\emph{values} = per-story vector, \emph{peak} = peak drift, \emph{one} =
one designated story's drift. \emph{comp} = held-out compliance accuracy
(balanced accuracy in parentheses; the majority baseline is 0.62 on the
forward stream and 0.98 on the post-intervention stream, where the
readout is uninformative; \arm{O}'s forward compliance accuracy exceeds
the majority baseline, exact binomial $p<0.001$). A parse failure scores as a miss; every parse
rate is 1.00. Ranking cells reproduce Table~\ref{tab:d1}.}
\label{tab:readout}\vspace{1.2ex}\scriptsize\setlength{\tabcolsep}{2.6pt}
\begin{tabular}{llccccccc}
\toprule
arm (trained on) & axis & rank & gov & val-argmax & values & peak & one & comp (bal.) \\
\midrule
\arm{P} (ranking) & forward & 0.73/0.89 & 0.74/0.87 & 0.47/0.57 & 0.87/0.93 & 0.69/0.79 & 1.00/1.00 & 0.57/0.51 (0.59/0.59) \\
 & extrapolation & 0.76/0.84 & 0.80/0.79 & 0.47/0.56 & 0.64/0.92 & 0.54/0.70 & 1.00/0.98 & 0.60/0.65 (0.60/0.66) \\
 & two-step & 0.58/0.73 & 0.43/0.69 & 0.39/0.31 & 1.07/0.92 & 1.77/1.98 & 2.81/2.64 & 0.97/0.79 (0.66/0.40) \\
 & post-intervention & 0.71/0.71 & 0.74/0.71 & 0.24/0.37 & 1.12/1.14 & 0.88/1.34 & 2.95/2.29 & 0.90/0.63 (0.79/0.49) \\
\addlinespace[2pt]
\arm{G} (governing story) & forward & 0.68/0.85 & 0.77/0.88 & 0.55/0.61 & 0.97/0.92 & 0.67/0.85 & 0.98/0.99 & 0.46/0.37 (0.48/0.41) \\
 & extrapolation & 0.71/0.82 & 0.81/0.82 & 0.55/0.61 & 0.93/0.93 & 0.57/0.71 & 0.99/0.99 & 0.57/0.50 (0.58/0.50) \\
 & two-step & 0.62/0.74 & 0.69/0.74 & 0.45/0.50 & 2.60/1.44 & 1.35/1.54 & 2.02/2.63 & 0.90/0.54 (0.62/0.60) \\
 & post-intervention & 0.67/0.70 & 0.70/0.72 & 0.37/0.47 & 3.02/1.72 & 0.87/1.21 & 1.88/2.35 & 0.63/0.41 (0.81/0.54) \\
\addlinespace[2pt]
\arm{V} (per-story values) & forward & 0.61/0.52 & 0.65/0.62 & 0.82/0.89 & 0.38/0.29 & 0.50/0.62 & 0.58/0.40 & 0.63/0.73 (0.62/0.73) \\
 & extrapolation & 0.64/0.55 & 0.70/0.68 & 0.79/0.91 & 0.36/0.28 & 0.48/0.63 & 0.49/0.48 & 0.62/0.70 (0.63/0.70) \\
 & two-step & 0.40/0.25 & 0.52/0.62 & 0.61/0.73 & 0.35/0.24 & 0.41/0.34 & 0.46/0.29 & 0.97/0.91 (0.50/0.79) \\
 & post-intervention & 0.31/0.30 & 0.41/0.64 & 0.69/0.76 & 0.33/0.25 & 0.46/0.52 & 0.53/0.34 & 0.97/0.92 (0.66/0.80) \\
\addlinespace[2pt]
\arm{K} (one story's value) & forward & 0.61/0.56 & 0.63/0.65 & 0.59/0.67 & 0.55/0.43 & 0.51/0.56 & 0.61/0.39 & 0.55/0.60 (0.59/0.65) \\
 & extrapolation & 0.61/0.61 & 0.69/0.66 & 0.68/0.72 & 0.50/0.42 & 0.49/0.49 & 0.53/0.30 & 0.66/0.69 (0.66/0.69) \\
 & two-step & 0.33/0.45 & 0.22/0.39 & 0.45/0.48 & 0.46/0.39 & 0.44/0.26 & 0.34/0.24 & 0.75/0.63 (0.87/0.81) \\
 & post-intervention & 0.35/0.35 & 0.25/0.38 & 0.33/0.35 & 0.40/0.40 & 0.43/0.33 & 0.30/0.34 & 0.79/0.65 (0.89/0.82) \\
\addlinespace[2pt]
\arm{A} (peak value) & forward & 0.47/0.51 & 0.50/0.65 & 0.47/0.51 & 0.79/0.61 & 0.48/0.28 & 0.93/0.69 & 0.63/0.59 (0.67/0.67) \\
 & extrapolation & 0.46/0.51 & 0.53/0.72 & 0.51/0.50 & 0.83/0.58 & 0.37/0.24 & 0.86/0.74 & 0.70/0.70 (0.70/0.71) \\
 & two-step & 0.42/0.30 & 0.37/0.49 & 0.43/0.23 & 1.12/0.56 & 1.17/0.20 & 1.83/0.80 & 0.95/0.86 (0.49/0.93) \\
 & post-intervention & 0.42/0.27 & 0.39/0.45 & 0.20/0.16 & 1.41/0.65 & 0.86/0.19 & 2.09/0.57 & 0.83/0.79 (0.91/0.89) \\
\addlinespace[2pt]
\arm{O} (compliance) & forward & 0.33/0.15 & 0.33/0.25 & 0.27/0.25 & 0.79/0.79 & 0.68/0.74 & 1.00/0.79 & 0.79/0.79 (0.75/0.72) \\
 & extrapolation & 0.34/0.16 & 0.34/0.30 & 0.23/0.24 & 0.84/0.86 & 0.54/0.82 & 1.00/0.79 & 0.71/0.72 (0.71/0.72) \\
 & two-step & 0.27/0.21 & 0.21/0.17 & 0.24/0.21 & 1.15/0.81 & 0.82/0.55 & 2.34/1.23 & 0.98/0.98 (0.50/0.50) \\
 & post-intervention & 0.19/0.12 & 0.16/0.13 & 0.19/0.13 & 1.64/0.85 & 0.67/0.56 & 2.77/0.87 & 0.98/0.98 (0.50/0.50) \\
\addlinespace[2pt]
\arm{S} (shuffled placebo) & forward & 0.35/0.24 & 0.40/0.27 & 0.13/0.23 & 0.88/0.79 & 0.70/0.78 & 1.70/0.79 & 0.38/0.62 (0.50/0.50) \\
 & extrapolation & 0.40/0.20 & 0.44/0.29 & 0.23/0.24 & 0.92/0.89 & 0.63/0.85 & 2.12/0.83 & 0.49/0.51 (0.50/0.50) \\
 & two-step & 0.27/0.15 & 0.09/0.09 & 0.24/0.21 & 2.77/1.15 & 4.84/0.51 & 21.39/3.75 & 0.02/0.98 (0.50/0.50) \\
 & post-intervention & 0.10/0.04 & 0.08/0.15 & 0.14/0.13 & 2.95/1.40 & 3.70/0.58 & 10.34/1.49 & 0.02/0.98 (0.50/0.50) \\
\addlinespace[2pt]
untrained base & forward & 0.30/0.25 & 0.40/0.23 & 0.35/0.36 & 1.92/0.73 & 0.72/0.61 & 1.78/0.85 & 0.37/0.38 (0.49/0.50) \\
 & extrapolation & 0.29/0.31 & 0.44/0.33 & 0.40/0.36 & 1.45/0.86 & 0.62/0.64 & 2.16/0.96 & 0.47/0.49 (0.49/0.50) \\
 & two-step & 0.28/0.13 & 0.17/0.07 & 0.31/0.19 & 5.17/2.08 & 4.22/2.36 & 26.43/10.43 & 0.02/0.02 (0.50/0.50) \\
 & post-intervention & 0.16/0.08 & 0.13/0.08 & 0.25/0.11 & 6.68/2.27 & 3.39/1.88 & 13.10/6.53 & 0.03/0.02 (0.50/0.50) \\
\bottomrule
\end{tabular}
\end{table*}
\begin{table*}[t]\centering
\caption{Four held-out tier-1 forward instances with every arm's verbatim
ranking (no worked examples; the governing story is the first entry).
\arm{G} was trained to name the governing story and returns a short list
that the shared parser completes; \arm{A}, \arm{O}, \arm{K}, and \arm{V}
were asked the same ranking question, and the table shows the ranking
each emitted.
Cases 1 and 3 are instances where the surface cue (the story with the
lightest columns) and the oracle disagree; case 4 is a near-tie that the
ranking-trained arm inverts.}
\label{tab:cases}\vspace{1.2ex}\scriptsize\setlength{\tabcolsep}{2.5pt}
\begin{tabular}{llllllllll}
\toprule
case & oracle & \arm{P} & \arm{G} & \arm{V} & \arm{K} & \arm{A} & \arm{O} & \arm{S} & base \\
\midrule
1 & [5,3,2,4,1] & [5,3,2,4,1] & [5,3] & [3,5,1,2,4] & [3,5,4,2,1] & [3,1,2,4,5] & [1,2,3,4,5] & [3,5,1,2,4] & [3,1,2,4,5] \\
2 & [4,1,3,5,2] & [4,1,3,2,5] & [4,1] & [1,2,3,4,5] & [2,1,4,3,5] & [2,1,3,4,5] & [1,2,3,4,5] & [1,2,5,3,4] & [2,3,4,1,5] \\
3 & [3,4,2,1,5] & [3,2,4,5,1] & [3] & [2,3,4,5,1] & [2,3,4,5,1] & [2,1,3,4,5] & [1,2,3,4,5] & [1,2,5,3,4] & [1,2,3,4,5] \\
4 & [4,3,5,6,2,1] & [3,4,5,6,2,1] & [3,4,5,6,3,4] & [1,2,3,4,5,6] & [3,4,5,6,2,1] & [1,2,3,4,5,6] & [1,2,3,4,5,6] & [1,2,3,4,5,6] & [1,2,3,4,5,6] \\
\bottomrule
\end{tabular}\par\vspace{3pt}
\parbox{\textwidth}{\footnotesize Case 1: story 3 has the lightest columns but is braced; story 5 is the only unbraced story. Case 2: story 4 is the only unbraced story (the surface cue and the oracle agree). Case 3: only the ranking-trained arm localizes the governing story. Case 4: a case the ranking-trained arm gets wrong (top-2 drift margin 2.3\%).}
\end{table*}

\begin{table*}[t]\centering
\caption{GRPO runs: start checkpoint, prompt pool after difficulty
filtering, group size $g$, optimizer steps, learning rate, and KL
coefficient $\beta$, and per-device batch $\times$ accumulation steps,
read from each run's saved training arguments.}
\label{tab:grpo}\vspace{1.2ex}\footnotesize
\begin{tabular}{llrrrlll}
\toprule
run (\S) & start & pool & $g$ & steps & lr & $\beta$ & batch \\
\midrule
round-1 judge (\S\ref{sec:verifier}) & \arm{P} & 2000 & 8 & 400 & $10^{-6}$ & 0 & 16$\times$2 \\
round-2 judge (\S\ref{sec:verifier}) & \arm{P} & 1707 & 16 & 600 & $10^{-6}$ & 0.01 & 16$\times$2 \\
matched, lr $10^{-5}$ (\S\ref{sec:verifier}) & \arm{P} & 1707 & 16 & 600 & $10^{-5}$ & 0.01 & 16$\times$2 \\
matched, lr $3{\times}10^{-5}$ (\S\ref{sec:verifier}) & \arm{P} & 1707 & 16 & 600 & $3{\times}10^{-5}$ & 0.01 & 16$\times$2; 4$\times$8 after step 500 \\
matched, lr $10^{-4}$ (\S\ref{sec:verifier}) & \arm{P} & 1707 & 16 & 600 & $10^{-4}$ & 0.01 & 4$\times$8 \\
best-practice $g{=}50$ (\S\ref{sec:verifier}) & SFT-hard & 344 & 50 & 400 & $10^{-6}$ & 0 & 10$\times$2 \\
lower-start $g{=}50$ (\S\ref{sec:verifier}) & \arm{P} & 597 & 50 & 400 & $10^{-6}$ & 0 & 10$\times$2 \\
lr probe (\S\ref{sec:verifier}) & SFT-hard & 344 & 50 & 400 & $10^{-5}$ & 0 & 10$\times$5 \\
from scratch (\S\ref{sec:verifier}) & base & 4000 & 16 & 800 & $10^{-6}$ & 0 & 16$\times$2 \\
action-format (\S\ref{sec:routing}) & SFT-hard & 1232 & 16 & 500 & $10^{-6}$ & 0.01 & 16$\times$2 \\
nothink-RL, Y (\S\ref{sec:think}) & SFT-hard & 696 & 16 & 1200 & $10^{-6}$ & 0 & 16$\times$2 \\
think-RL, X (\S\ref{sec:think}) & bridge & 759 & 14 & 600 & $10^{-6}$ & 0 & 1$\times$8 \\
\bottomrule
\end{tabular}
\end{table*}
\begin{table*}[t]\centering
\caption{Matched-recipe cells (\S\ref{sec:verifier}). Tier-2 localization on
the four axes ($n{=}150$ per axis, seed 42, single checkpoint), tier-1 forward
retention where benched, and exact McNemar tests on the shared tier-2 forward
stream against the start checkpoint and against GRPO round-2 (discordant pairs
$a{:}b$, $p$). All rows start from the tier-1 \arm{P} arm except the lr probe,
which starts from the dense-SFT checkpoint. The two $10^{-5}$ cells differ on 11:11 discordant pairs (SFT:GRPO, $p{=}1.00$). Pooled $n{=}600$
contrasts for the matched cells are in Table~\ref{tab:d7}.}
\label{tab:matched}\vspace{1.2ex}\footnotesize\setlength{\tabcolsep}{3.5pt}
\begin{tabular}{lccccccc}
\toprule
cell & fwd & extrap & two-step & post-int. & tier-1 fwd & vs.\ start & vs.\ GRPO r2 \\
\midrule
start (\arm{P} arm) & 0.660 & 0.575 & 0.593 & 0.553 & 0.887 & -- & 5:9, 0.42 \\
GRPO round-2 ($10^{-6}$, 600 steps, $g{=}16$) & 0.687 & 0.588 & 0.593 & 0.633 & -- & 9:5, 0.42 & -- \\
dense SFT ($10^{-4}$, 940 steps, 2 epochs) & 0.747 & 0.713 & 0.707 & 0.693 & -- & 22:9, 0.03 & 20:11, 0.15 \\
matched SFT ($10^{-6}$, 600 steps) & 0.667 & 0.588 & 0.593 & 0.567 & 0.893 & 1:0, 1.00 & 5:8, 0.58 \\
matched SFT ($10^{-5}$, 600 steps) & 0.760 & 0.625 & 0.613 & 0.660 & 0.880 & 16:1, 0.0003 & 13:2, 0.007 \\
matched SFT ($3{\times}10^{-5}$, 600 steps) & 0.767 & 0.637 & 0.640 & 0.640 & 0.873 & 22:6, 0.004 & 16:4, 0.01 \\
matched SFT ($10^{-4}$, 600 steps) & 0.687 & 0.637 & 0.687 & 0.667 & 0.827 & 21:17, 0.63 & 18:18, 1.00 \\
matched GRPO ($10^{-5}$, 600 steps, $g{=}16$) & 0.760 & 0.613 & 0.607 & 0.667 & 0.720 & 24:9, 0.01 & 18:7, 0.04 \\
matched GRPO ($3{\times}10^{-5}$, 600 steps, $g{=}16$) & 0.627 & 0.550 & 0.527 & 0.600 & 0.533 & 20:25, 0.55 & 16:25, 0.21 \\
matched GRPO ($10^{-4}$, 600 steps, $g{=}16$) & 0.487 & 0.487 & 0.420 & 0.367 & 0.407 & 11:37, 0.0002 & 9:39, 2e-5 \\
GRPO lr probe ($10^{-5}$, from dense SFT) & 0.680 & 0.688 & 0.607 & 0.613 & -- & 18:15, 0.73 & 17:18, 1.00 \\
\bottomrule\end{tabular}
\par\vspace{2pt}\parbox{\linewidth}{\footnotesize Fresh $n{=}600$ streams, per axis (forward/two-step/post-intervention; the pooled contrasts of Table~\ref{tab:d7} use the same files): start 0.628/0.643/0.595; dense SFT 0.757/0.695/0.695; GRPO round-2 0.655/0.620/0.612; matched SFT $10^{-6}$ 0.653/0.643/0.615; matched SFT $10^{-5}$ 0.700/0.640/0.657; matched SFT $3{\times}10^{-5}$ 0.720/0.640/0.655; matched SFT $10^{-4}$ 0.667/0.663/0.657; matched GRPO $10^{-5}$ 0.697/0.588/0.635; matched GRPO $3{\times}10^{-5}$ 0.613/0.530/0.568; matched GRPO $10^{-4}$ 0.485/0.438/0.380.}
\end{table*}

\end{document}